# Interactive Decision Tree Creation and Enhancement with Complete Visualization for Explainable Modeling

Boris Kovalerchuk, Andrew Dunn, Alex Worland, Sridevi Wagle
Department of Computer Science, Central Washington University, USA
Boris.Kovalerchuk@cwu.edu, Andrew.Dunn@cwu.edu, Alex.Worland@cwu.edu, Sridevi.Wagle@cwu.edu

**ABSTRACT**

To increase the interpretability and prediction accuracy of the Machine Learning (ML) models, visualization of ML models is a key part of the ML process. Decision Trees (DTs) are essential in machine learning (ML) because they are used to understand many black box ML models including Deep Learning models. In this research, two new methods for creation and enhancement with complete visualizing Decision Trees as understandable models are suggested. These methods use two versions of General Line Coordinates (GLC): Bended Coordinates (BC) and Shifted Paired Coordinates (SPC). The Bended Coordinates are a set of line coordinates, where each coordinate is bended in a threshold point of the respective DT node. In SPC, each n-D point is visualized in a set of shifted pairs of 2-D Cartesian coordinates as a directed graph. These new methods expand and complement the capabilities of existing methods to visualize DT models more completely. These capabilities allow us to observe and analyze: (1) relations between attributes, (2) individual cases relative to the DT structure, (3) data flow in the DT, (4) sensitivity of each split threshold in the DT nodes, and (5) density of cases in parts of the n-D space. These features are critical for DT models' performance evaluation and improvement by domain experts and end users as they help to prevent overgeneralization and overfitting of the models. The advantages of this methodology are illustrated in the case studies on benchmark real-world datasets. The paper also demonstrates how to generalize them for decision tree visualizations in different General Line Coordinates.

## 1. INTRODUCTION

Advanced methods are needed for the evaluation and improvement of Machine Learning (ML) models, including their **interpretability** and **prediction accuracy** [3]. Visualization plays an important role in this [4]. Decision trees (DTs) are among a few machine learning methods, which are commonly recognized as interpretable models. Decision trees are also used to interpolate much more complex black box machine learning models. However, there are several difficulties known with the decision trees. They may not be stable and there are multiple different decision trees that have very similar accuracy and interpretation power. Therefore, efficient visualization of the decision trees can help significantly in resolving these issues, which is the focus of this work.

While decision trees are important as was outlined above, the DT topic is a part of a more general issue of visual knowledge discovery (VKD). The goal of the visual knowledge discovery is tight integration of artificial intelligence and machine learning methods with advanced visualization methods for efficient knowledge discovery [3, 31]. One of the focuses of this integration is incorporating visual methods to the core of the model discovery, enhancement, and evaluation process to make this process visual as much as possible.

In the current practice visualization often is a separate stage for the models already built by analytical methods like decision trees. Several important characteristics of the proposed approach demonstrate that it is leaning to the core of the AI/ML modeling process with visual means. First it allows building decision trees interactively visually without using analytical machine learning algorithms. In includes interactive changing of thresholds. Next, it allows losslessly visualize all training and validation data along with the DT model. This is a way to see all borderline cases to be able to adjust decision tree interactively to make it more robust. Moreover, it opens an opportunity for discovering visually models, which are more general than decision trees as it is shown in Section 5.

We offer new visualization methods BC-DT and SPC-DT. The concept of **General Line Coordinates** (GLC) [3,9] is behind both methods. Specifically, BC-DT and SPC-DT methods rely on new **Bended Coordinates** (**BC**) and the **Shifted Paired Coordinates** (**SPC**), respectively. We also **generalize** these methods to several other GLCs. The Bended Coordinates are line coordinates that are bended in a specific point. In the BC-DT, this point is a threshold assigned to the given node of the DT. As the name indicates, SPC is set of shifted pairs of Cartesian coordinates. In SPC, a directed graph (digraph) visualizes an n-D point **x** in 2-D with nodes formed by consecutive pairs $(x_i, x_{i+1})$ of values of coordinates of **x,** connected by directed edges.

These methods expand capabilities available in the traditional visualization of DT models by visualizing: (1) relations between attributes, (2) individual cases relative to the DT, (3) data flow in the DT, (4) how tight is each split threshold in the DT nodes, and (5) density of cases in parts of the n-D space. This information has a significant value for domain experts in evaluating and improving DT models.

In 2011, [8] reported on about 200 different DT representations; today, Treevis.net displays over 300. However, typically they are not detailed enough to represent the machine learning models. For example, the split thresholds in the DT machine learning models are not visible using a Treemap [11] visualization, which is particularly effective for displaying big, complicated trees. Some DT visualizations for ML present the number of cases in each node marginal distribution of the cases [5-7].

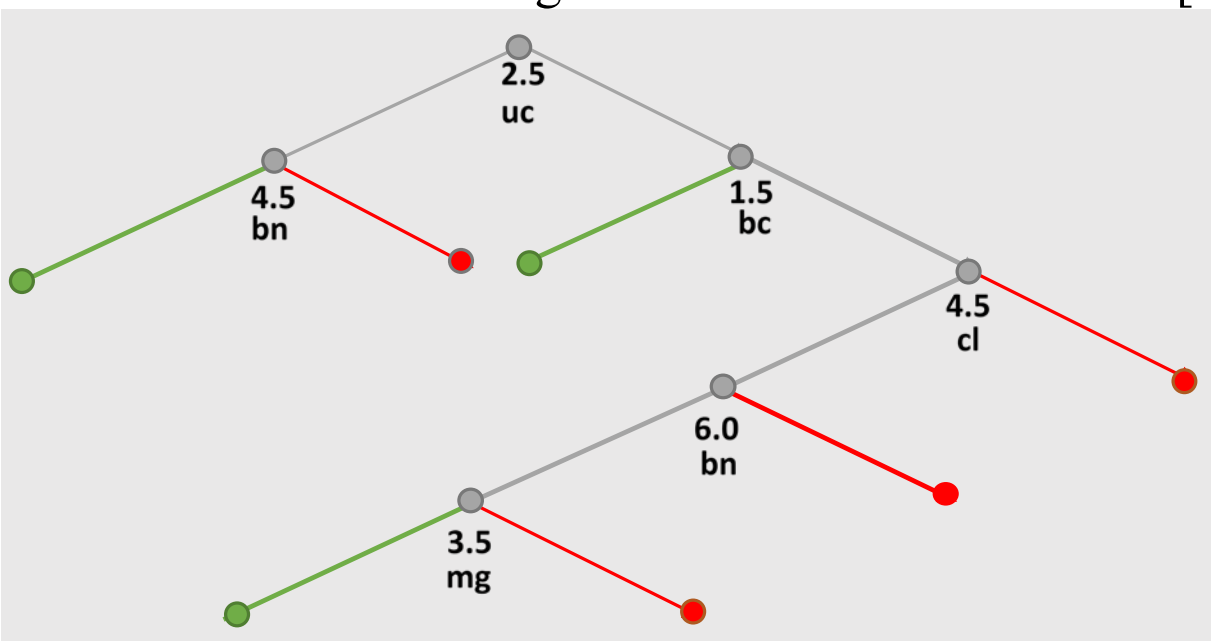

Fig. 1. Traditional decision tree representation of Wisconsin Breast Cancer (WBC) data. The benign class is denoted by green edges and nodes, while the malignant class is denoted by red edges and nodes. Cases that have not yet reached the DT level are shown by gray edges and nodes.

The decision tree's common picture, as displayed in Fig. 1, merely shows the DT's overall structure. Using the representation in Fig. 1, a new n-D point may be predicted by following a classification prediction from the root to the leaf. This does not adequately convey the accuracy of this forecast, though. The DT visualization shown in Fig. 2 [2] is also limited not showing individual cases. We need additional knowledge to answer such questions, which is nor in such visualizations.

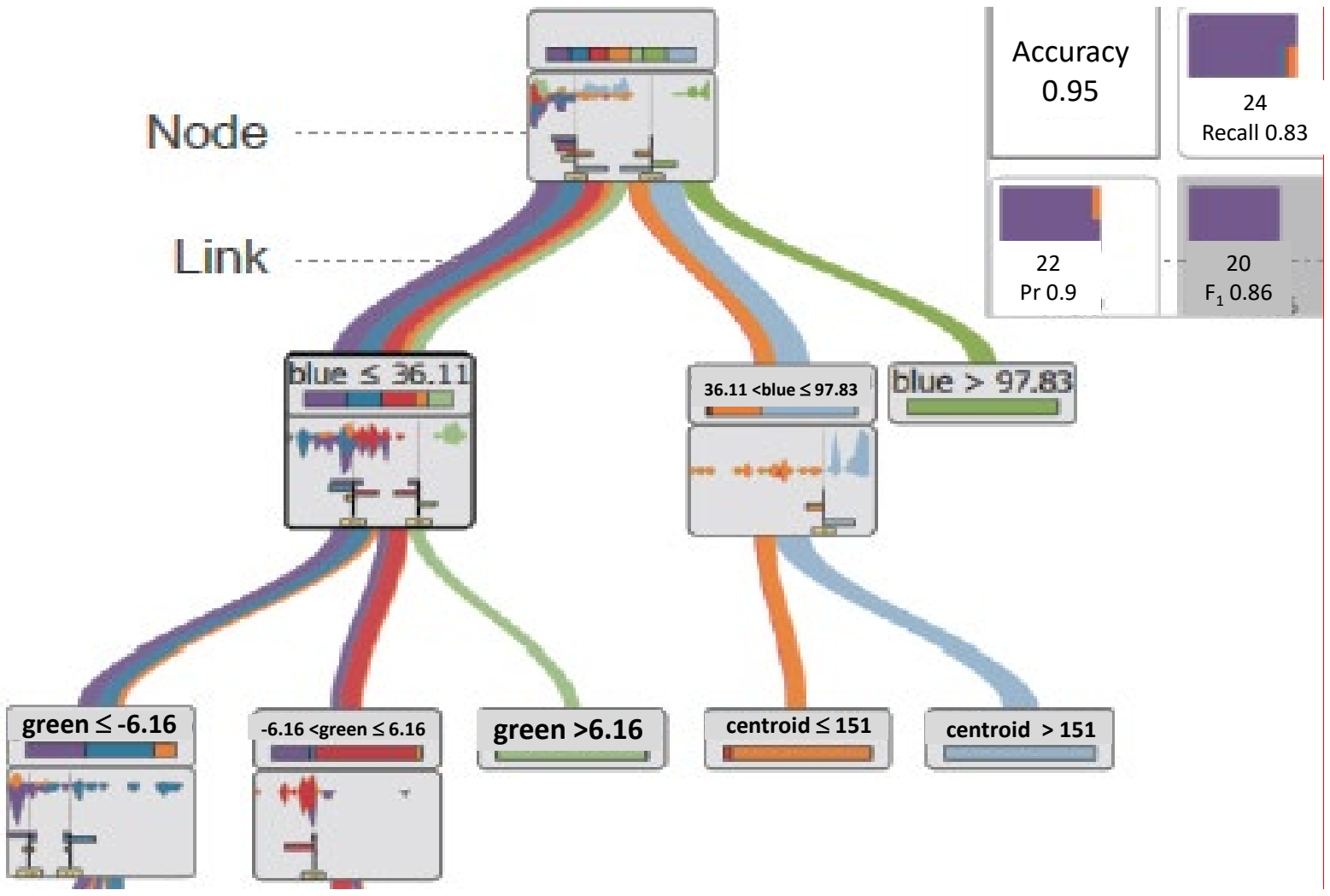

Fig. 2. DT visualization as streams [2].

The fact that DT node thresholds are not unique is well known [12,13]. We can maintain the same DT prediction on training and/or validation data by changing thresholds within certain constraints. The fluctuation of the thresholds makes prediction of new cases close to the threshold instable, which can change the predicted class dramatically. The study of these scenarios is difficult with current decision tree visualization techniques.

Tight thresholds [12,13] are a concern when cases from opposing classes overlap and are near to the threshold. Setting up the proper criteria automatically is challenging. As we just examined above about new borderline cases, a slight change in the threshold can result in classification changes for many training, validation, and new cases, making the prediction incorrect. Thus, for the tight borders, many training and validation cases are borderline cases that needs to be analyzed.

The ability to apply the domain knowledge provided by the end users makes it possible to adjust thresholds via interaction with graphics, which is a big advantage. The current DT visualizations are insufficient in this regard. Another well-known issue with DTs is **overfitting**, which necessitates branch shortening to prune DTs. Although the number of thresholds is reduced, DT accuracy is decreased. It is challenging to strike a balance between the accuracy and size of the tree. By including the end users' domain expertise, the visualization can aid in creating the right balance.

Additionally, current visualizations do not enable displaying all individual training cases along in the DT, nor do they support displaying all individual validation cases in the DT to compare their representation in the DT. For instance, Figs. 1 and 2 make it difficult to perceive borderline scenarios. Such decision tree's capacity for in-depth examination and improvement is constrained by these flaws.

The suggested methods enable the visualization of all training, validation, and test data in the DT, individually or collectively. Since this visualization is lossless [3,9], it will reveal relations concerning training, validation and test data and will assess the stability and **trustworthiness** of the decision tree. It also makes it possible to see data **outliers**.

Convincing the user that the DT prediction can be trusted is the challenge for the decision tree model. Unfortunately, the user's ability to be confident is constrained by the current visualization techniques. To solve this problem, the suggested method enables the decision tree to trace any new case in relation to the training and validation data. Also, this aids in justifying or modifying the DT thresholds for borderline cases.

Overfitting is a prominent issue with decision trees, although **overgeneralization** is less visible but still crucial [10]. Consider attribute $x_1$ in the range [0,10] and a branch of the DT, which states that if $x_1 \leq 2.5$, then the case belongs to class 1. Next, no case of class 1 has $x_i$ that is less than 1.5 and all training cases are within that range [1.5, 2.5]. The decision three overgeneralized cases of class 1 to the interval [0, 2.5]. The suggested visualization technique makes it possible to detect overgeneralization and adjust the range to its true bounds.

This paper is organized as follows. Section 2 defines methods, sections 3 and 4 describe case studies on the benchmark datasets, Section 5 presents a discussion and generalizations of the methods and Section 6 summarizes the chapter.

## 2. METHODS

### 2.1. Traditional methods vs. coordinate-based methods

Below the **Bended Coordinates Decision Tree** (**BC-DT**) method is explained, using the ID3 Decision Tree shown in Fig. 1 above in the traditional visualization, trained on 349 cases of for Wisconsin Breast Cancer (WBC) data [1]. These cases are about 50% of all WBC Data. Fig. 3 and Table 1 describe this DT and its performance. While WBC data contain nine attributes, this decision tree uses only five attributes, where the attribute bnuclei (bn) is used twice with two different splits. Thus, we have 5-D data in this DT that we represent as a 6-D point with bnuclei repeated.

The root of this tree is based on the attribute ucellsize (uc), the next node is based on the attribute bnuclei. The proposed methods focus on the abilities to see how the **individual cases traverse** through the DT, and how **close** they to the **thresholds** associated with the DT nodes. The traditional DT visualization, in Fig. 1, has no such capability. This goal was partially reached in [22] by showing n-D data in parallel coordinates side-by-side with a traditional DT (see Fig. 4a), but not in the tree itself. It is interactive and a user can click on any node in the tree to see data reaching the node in the parallel coordinates in a separate coordinated plot.

- ucellsize < 2.5
- bnuclei < 4.5 then class = **benign** (100.00 % of 200 cases)
- bnuclei ≥ 4.5 then class = **malignant** (66.67 % of 6 cases)
- ucellsize ≥ 2.5
- bchromatin < 1.5 then class = **benign** (87.50 % of 8 cases)
- bchromatin ≥ 1.5
    - clump < 4.5
    - bnuclei < 6.0
        - mgadhesion < 3.5 then class = **benign** (100.00 % of 5 cases)
        - mgadhesion ≥ 3.5 then class = **malignant** (66.67 % of 6 cases)
    - bnuclei ≥ 6.0 then class = **malignant** (100.00 % of 8 cases)
    - clump ≥ 4.5 then class = **malignant** (93.97 % of 116 cases)

Fig. 3. ID3 Decision Tree for WBC data.

Table 1. ID3 DT performance.

| Error rate | | | 0.0716 | | | |
|---|---|---|---|---|---|---|
| Values prediction | | | Confusion matrix | | | |
| Value | Recall | 1-Precision | | benign | malignant | Sum |
| benign | 0.9056 | 0.0194 | benign | 202 | 21 | 223 |
| malignant | 0.9683 | 0.1469 | malignant | 4 | 122 | 126 |
| | | | Sum | 206 | 143 | 349 |

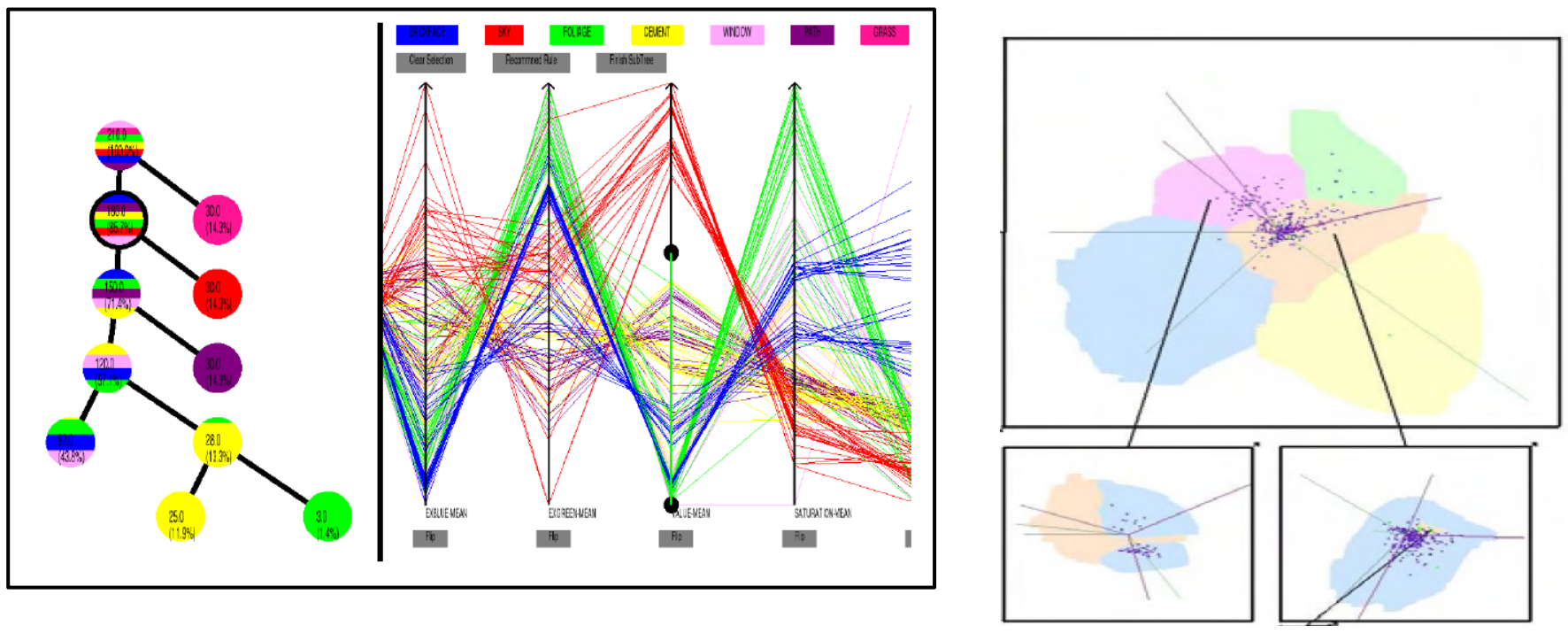

(a) n-D data in Parallel Coordinates side-by-side with a traditional DT [14].

(b) Star Coordinates space partitioning to build a tree structure [15].

Fig. 4. Alternative coordinate-based methods.

This is an example of the use of **coordinate-based methods** to visualize DTs, because representation of n-D data in parallel coordinates is one of the coordinate-based methods. All these methods belong to the class of lossless **General Line Coordinates** [3,9]. Another example of combining coordinate-based methods with DTs is presented in [15] using Kandogan' Star Coordinates [16] and Inselberg's Parallel Coordinates [17]. It allows a user interactively building a DT in these visualizations. To get a better data visualization a user can change Star Coordinates by moving each axis around by clicking on the endpoint of the selected axis and dragging it to the desired position.

This method produces traditional interpretable decision tree like in [22] when it uses parallel coordinates. However, when it uses star coordinates the situation is very different. The result is *not a traditional decision tree*. It is a hierarchical tree structure that partitions the star coordinate space. The points in this space are vector sums of the star coordinates. The application domain interpretation of these sums is not clear, in contrast with simple interpretable relations like $x_i > T_i$, where $T_i$ is a threshold value for individual coordinate/attribute $x_i$ captured in the traditional DTs. In [18] a method to interpret areas like found in [15] are converted to interpretable rules. Thus, an extra effort is needed to use models with Star Coordinates from [15] to derive interpretable models. In contrast with Fig. 4a, which shows a tradition DT visualization with underlying data side-by-side in parallel coordinates from [14], Fig. 4b illustrates the Star Coordinates space partitioning from [15] for a hierarchical tree structure.

In contrast Fig. 5 illustrates the method that we propose to combine DT and parallel coordinates in a single plot. It is the same DT as in Fig. 1. This visualization allows to trace individual cases in the decision three in contrast with Fig. 1. We will call this type of visualization as **Parallel Coordinate Decision Tree** (**PC-DT**). While this method allows to trace individual cases, it *does not show the traditional decision tree structure*. In the next section we present BC-DT method.

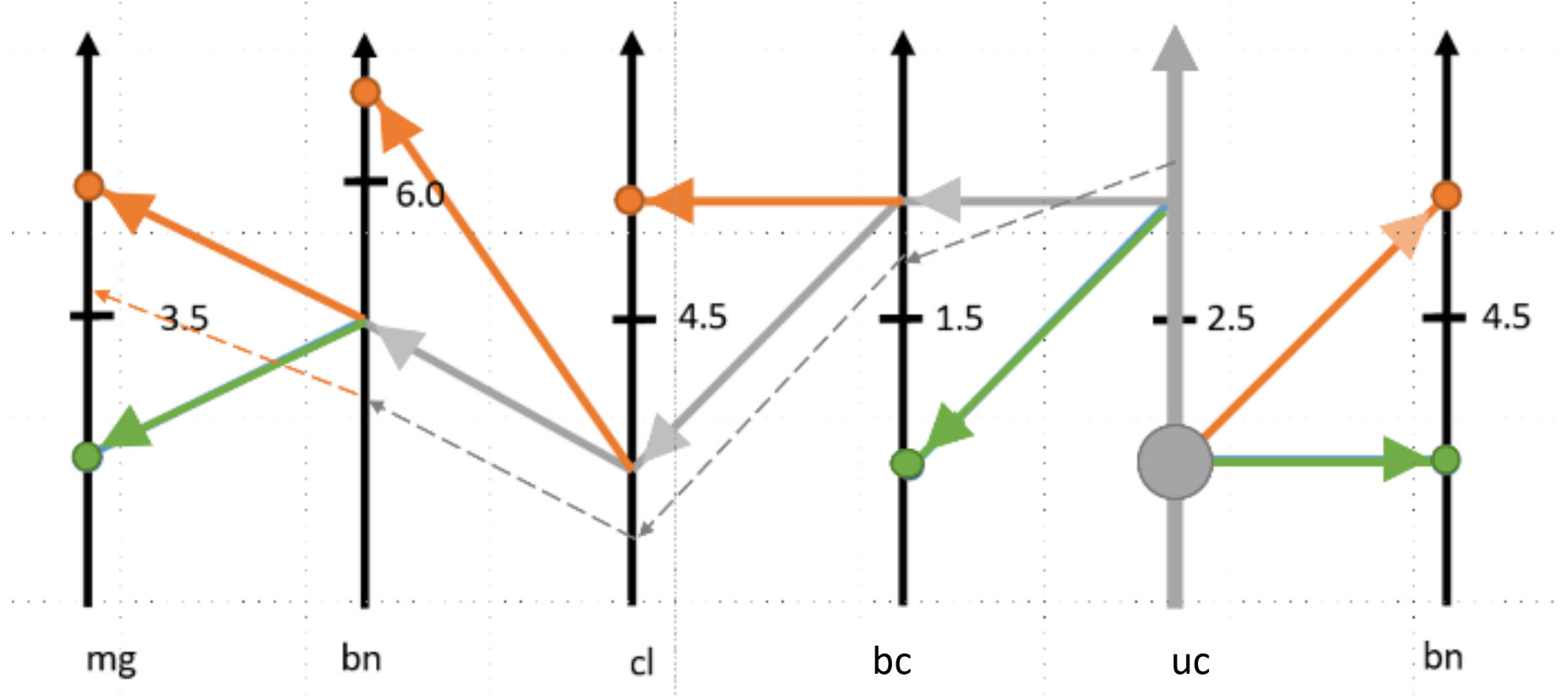

Fig. 5. DT visualized in Parallel Coordinates where coordinate uc is used as the root and the dotted line showing an individual case traced in the DT.

### 2.2. Bended Coordinates Decision Tree (BC-DT) method

The BC-DT method augments traditional DT visualization, by using **edges of DT as coordinates**, which are bended in the nodes at the threshold values. Thus, this is a **coordinate-based method**. Figs. 6 and 7 show two different ways how coordinates are bent at the threshold value.

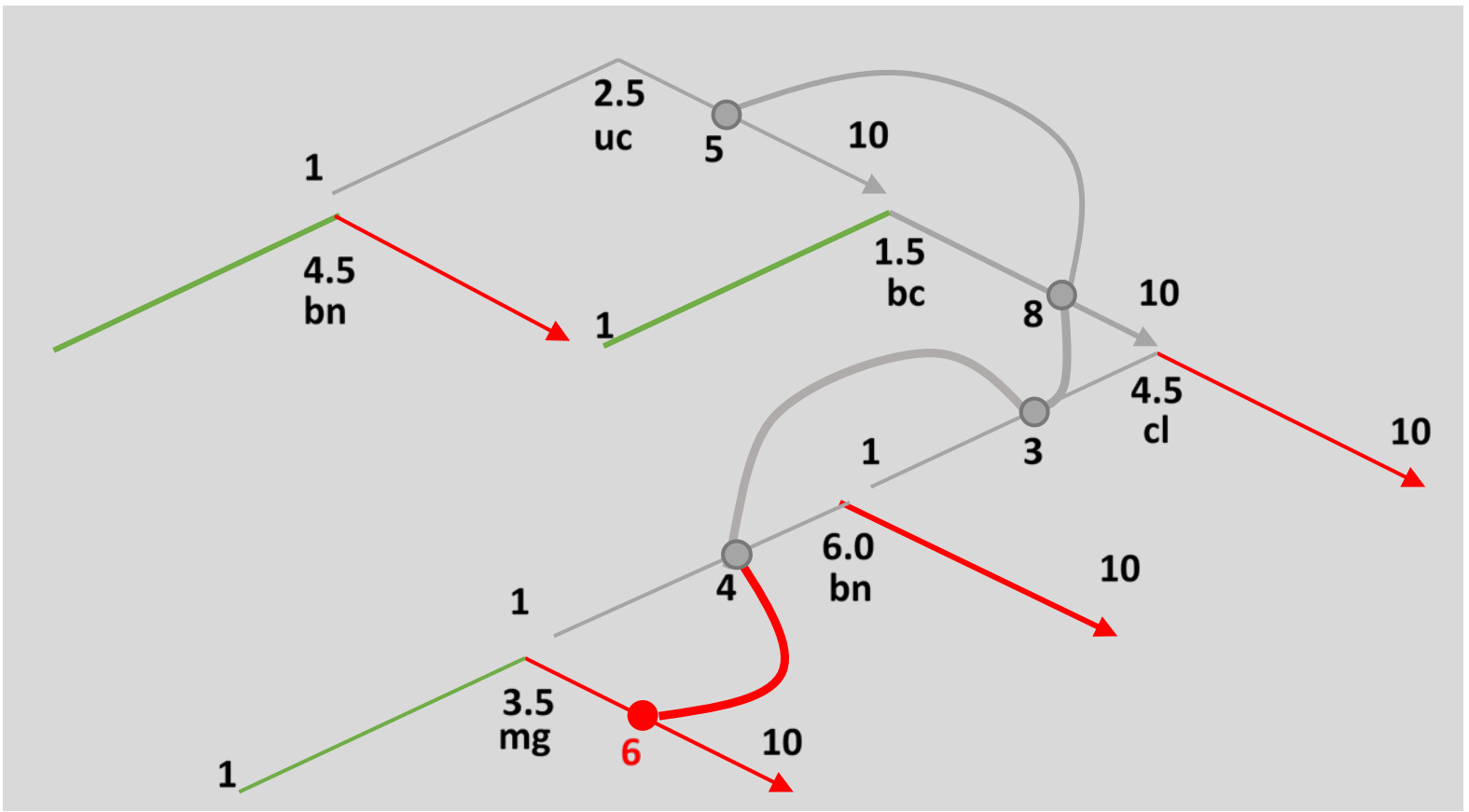

Fig. 6. Decision tree in BC-DT with edges serving as Bended Coordinates in disproportional scales for Wisconsin Breast Cancer Data. The curved solid line show 5-D point (uc, bc, cl, bn, mg) = (5, 8, 3, 4, 6) that reached a point mg=6 on this malignant edge of the DT.

For instance, the threshold $T$=2.5 on coordinate uc with interval of values [1,10] leads to the left interval [1, 2.5) and the right interval [2.5,10]. In Fig. 6, these two unequal intervals are visualized with **equal lengths**, i.e., forming **disproportional scale**. In Fig. 7a, the lengths of edges are 1.5 to 7.5 reflecting the location of the threshold $T$=2.5 in the interval [1,10]. In Fig. 7a, the dotted lines extend edges to make DT symmetric resembling the traditional DT visualization. Both Figs 6 and 7 show how a 5-D case is traversed in the DT classifies reaching malignant node indicated by the red arrow leading to it.

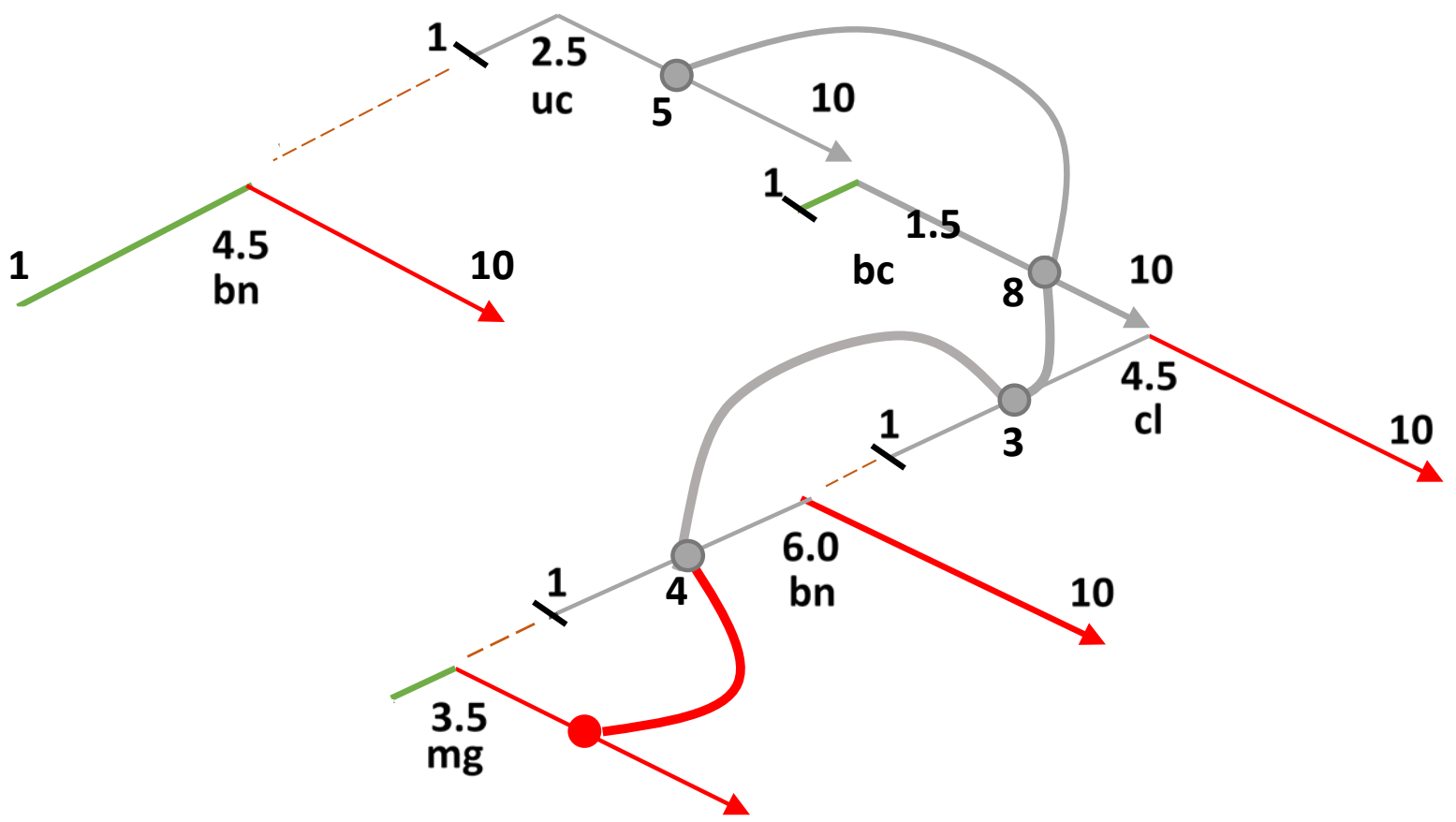

(a) 5-D point (5, 8, 3, 4, 6) in BC-DT that reaches the terminal malignant edge of the DT

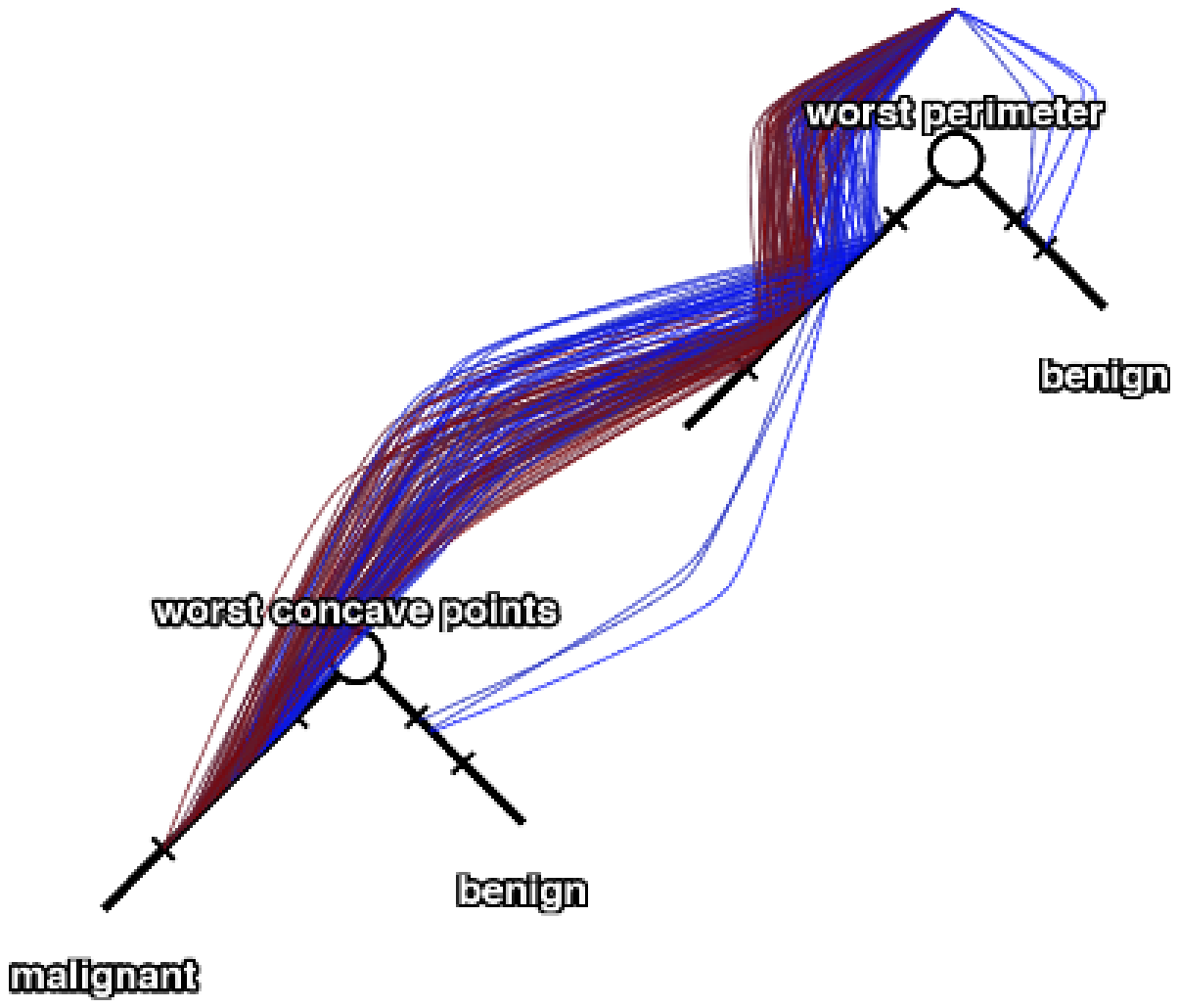

(b) Several cases in BC-DT in proportional scales

Fig. 7. BC-DT with proportional edges serving as Bended Coordinates.

Both Fig. 6 and Fig. 7a show 5-D point (uc, bc, cl, bn, mg) = (5, 8, 3, 4, 6). The distances from the thresholds to these values are different for different nodes. The shorter distances lead to less certainty in the classification of this 5-D point to the malignant class. The longer distances increase the confidence in the classification Relative to given thresholds. The unequal shoulders of the edges in Fig, 7a allow to see these differences undistorted in contrast with disproportional scaling in Fig. 6. Fig. 7b also shows DT with unequal shoulders of edges with undistorted distribution of many cases in this DT.

The BC-DT method expands and complements capabilities available in the traditional visualization of DT models by visualizing: (1) **individual cases** relative to the DT, (2) **detailed data flow** in the DT, and (3) the **tightness** of each split threshold in the DT nodes. This information is of a significant value to domain experts in evaluating and improving DT models. To the best of our knowledge, none of the existing DT visualization methods support all these capabilities. Section 4 provides case studies with BC-DT visualization on several benchmark datasets.

### 2.3. Shifted Paired Coordinates – Decision Tree (SPC-DT) method

The SPC-DT approach is also based on the concept of General Line Coordinates (GLC) [3]. The technique specifically uses the Shifted Paired Coordinates (SPC) representation. SPC, as its name suggests, is a collection of shifted pairs of Cartesian coordinates. An n-D point **x** is represented in 2-D by a directed graph (digraph) in SPC, where nodes are made up of consecutive pairs of **x**'s coordinate values ($x_i$, $x_{i+1}$) and edges connect them.

The SPC-DT technique enhances and completes the capabilities of the traditional DT model visualization by visualizing: (1) **relations** between attributes, (2) **individual cases** relative to the DT, (3) **detailed data flow** in the DT, (4) the **tightness** of each split threshold in the DT nodes, and (5) 2-D **density** of cases in areas of the n-D space. Experts in the field can greatly benefit from this knowledge in the process of evaluating and enhancing DT models. To the best of our knowledge, none of the DT visualization techniques now in use provide all of these capabilities.

Below, a conceptual explanation of the Shifted Paired Coordinates-Decision Tree (SPC-DT) method is followed by examples. In order to create a DT visualization from a given decision tree model, the SPC-DT process must go through the following major steps:

(1) Parsing the DT model.

(2) Pairing attributes.

(3) Building a set of paired Cartesian coordinates.

(4) Drawing each pair of coordinates in the default location

shifted one over another.

(5) Mapping a part of the decision tree to a respective pair of coordinates as rectangles based on the threshold values.

(6) Color rectangles according to classes.

(7) Drawing collection of n-D points as directed graphs in the SPC.

(8) Interactive modification of visualization. The options include:
(a) change a set of visualized n-D points,
(b) change the location of a coordinate pair,
(c) change colors of the classes,
(d) negate/flip coordinates,
(e) swap vertical and horizontal coordinates,
(f) condense points in a rectangle,
(g) show overall DT structure,
(h) adjust thresholds, and
(i) compute accuracy after adjusting thresholds.

The following example explains **Shifted Paired Coordinates** (**SPC**) in more detail. Take the 6-D point $\mathbf{x} = (x_1, x_2,..., x_6) = (1, 2, 3, 2, 5, 1)$. In SPC representation, a point in the first pair of coordinates $(x_1, x_2) = (1, 2)$ is visualized $(X_1, X_2)$. In the second set of coordinates $(X_3, X_4)$, the second pair $(x_3, x_4)=(3, 2)$ is visualized, and the third pair $(x_5, x_6)=(5, 1)$ is visualized $(X_5, X_6)$. In Fig. 8, a directed graph (1, 2) (3, 2) (5, 1) is formed by connecting these three points, which are shifted relative to one another to prevent overlap. Thus, shifted pairs of coordinates are generated.

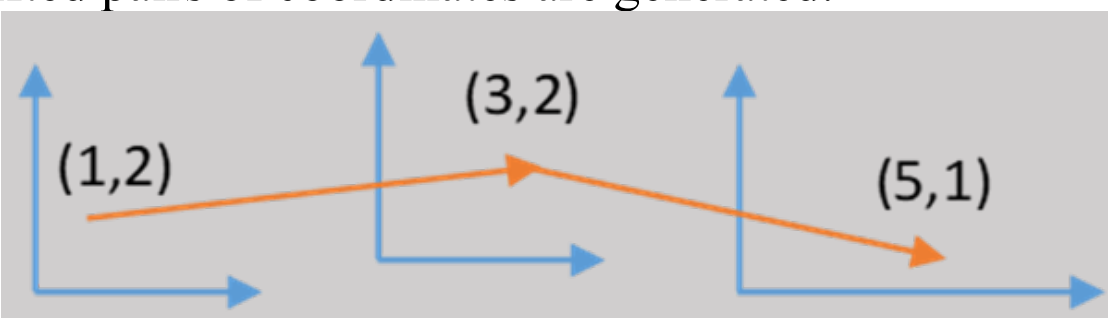

Fig. 8. 6-D point $x = (x_1,x_2,\ldots,x_6) = (1,2,3,2,5,1)$ in Shifted Paired Coordinates (SPC) visualization.

The decision tree in SPC-DT that was designed to visualize the WBC data is depicted in Fig. 9. Performance of the tree is shown in Table 2, and initial placements for each coordinate pair are shown for all WBC data in SPC-DT in Fig. 10. The gray areas on each paired coordinate space map denote areas where a case's class cannot be determined at the current DT level and must be followed to the next DT level to be evaluated further.

- ucellsize < 2.5000
    - bnuclei < 4.5000
        - clump < 6.5000 then classe = **begnin** (100.00 % of 407 examples)
        - clump >= 6.5000 then classe = **malignant** (60.00 % of 5 examples)
    - bnuclei >= 4.5000 then classe = **malignant** (52.94 % of 17 examples)
- ucellsize >= 2.5000
    - ucellsize < 4.5000
        - bnuclei < 2.5000
            - normnucl < 2.5000 then classe = **begnin** (100.00 % of 19 examples)
            - normnucl >= 2.5000 then classe = **begnin** (54.55 % of 11 examples)
        - bnuclei >= 2.5000
            - clump < 6.5000
                - bchromatin < 3.5000 then classe = **begnin** (63.64 % of 11 examples)
                - bchromatin >= 3.5000 then classe = **malignant** (83.33 % of 18 examples)
            - clump >= 6.5000
                - normnucl < 7.0000 then classe = **malignant** (100.00 % of 24 examples)
                - normnucl >= 7.0000 then classe = **malignant** (88.89 % of 9 examples)
    - ucellsize >= 4.5000
        - clump < 6.5000
            - bnuclei < 8.5000
                - bchromatin < 4.5000 then classe = **begnin** (50.00 % of 8 examples)
                - bchromatin >= 4.5000
                    - mgadhesion < 8.5000 then classe = **malignant** (100.00 % of 16 examples)
                    - mgadhesion >= 8.5000 then classe = **malignant** (75.00 % of 4 examples)
            - bnuclei >= 8.5000 then classe = **malignant** (100.00 % of 44 examples)
        - clump >= 6.5000 then classe = **malignant** (100.00 % of 106 examples)

. Fig. 9. ID3 Decision tree for WBC data.

TABLE 2. PERFORMANCE OF THE DECSION TREE FROM FIG. 9.

| Error rate | | | 0.0401 | | | |
|---|---|---|---|---|---|---|
| Values prediction | | | Confusion matrix | | | |
| Value | Recall | 1-Precision | | benign | malignant | Sum |
| benign | 0.9672 | 0.0285 | benign | 443 | 15 | 458 |
| malignant | 0.9461 | 0.0617 | malignant | 13 | 228 | 241 |
| | | | Sum | 456 | 243 | 699 |

To depict the various outcomes of instances in gray areas, several grayscale shades are used. There is just one destination plot for all cases that fall inside the same shade of gray. The DT has classified some cases as "malignant," while other cases have been classified as "benign," as indicated by the red and green areas. These cases terminate at the current level since a class has been determined.

The benefits of the SCP-DT visualization in Fig. 10 include:

1. relations between *2 attributes* inside of each plot,
2. relations between *4 attributes* using green and red arrows between two plots,
3. relations between *all 6 attributes* using green and red arrows between three plots and more attributes for larger dimensions,
4. the *tree structure* and data flow by arrows in the DT.

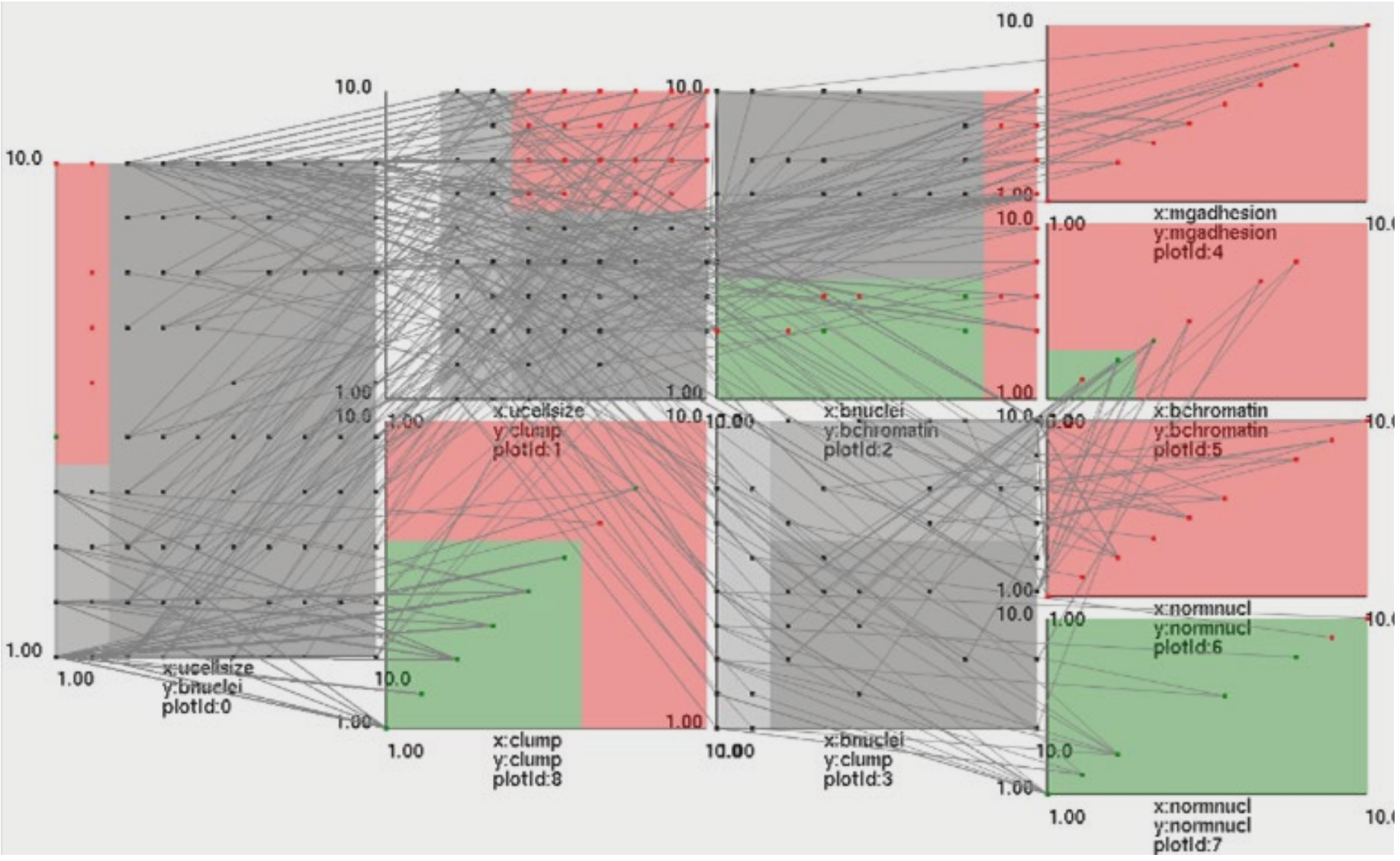

Fig. 10. WBC in default SPC-DT.

Many 10-D points in the SPC-DT are depicted in Fig. 10 as reaching the DT's terminal nodes. The proximity of such particular cases to each split threshold in the DT nodes can be seen on the graphs of those cases. Moreover, graphs for training instances make it possible to see how densely distributed the cases are in the n-D space. The number of instances in various areas of the space or the degree of class purity in the region can be displayed using color intensity (lighter/darker).

All these characteristics work together to boost trust in the DT model's ability to be understood and to predict accurately. Consider a DT where the actual training examples are all outside the threshold (border) area but a new case that needs to be forecasted is just on the border. Most likely, this DT model overgeneralized the training set, making the predictions excessively uncertain of such new case.

SPC-DT visualization aids domain experts in assessing the effectiveness, interpretation, overgeneralization, and overfitting of the DT model. The interactive SPC-DT can allow users to turn on and off "context" attributes, which are attributes that are not part of the tree but are a part of the input dataset. The tree root can be connected to these attributes. When such properties are disabled, they can be grayed out or made semitransparent so that the context may still be seen.

The SPC-DT design allows showing just subsets of cases when we need to visualize a large number of cases and classes. It can be "centers" of subsets, min or max cases of subsets with the remaining cases being gray or semitransparent  to decrease occlusion and simplify visual patterns. The SPC-DT design also allows showing the error rate for each dataset, branch of the tree, and the confusion matrix to aid the user in making confident predictions.

## 3. BC-DT Case Studies

### 3.1. Case study: seeds dataset

Below the Bended Coordinates DT approach (BC-DT) is presented as a case study with the seeds data. We show two versions for building the bended decision tree. The first one is based on the imported data and decision tree created outside of this program. In this case study the Tanagra system [22] was used as a source of the decision tree. In the second version, a user builds the DT interactively from scratch. A decision tree can be constructed either from scratch without any guidance or using a guidance form an existing DT, like the DT built in Tanagra [27] as a guiding prototype. A user can correct a prototype by changing thresholds and attributes or by removing some nodes or adding more nodes.

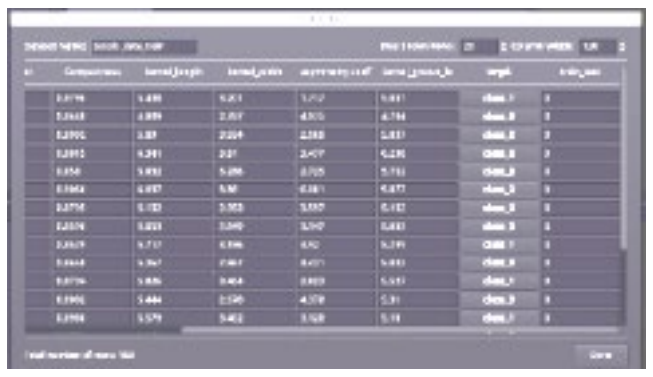

(a) Import seeds dataset.

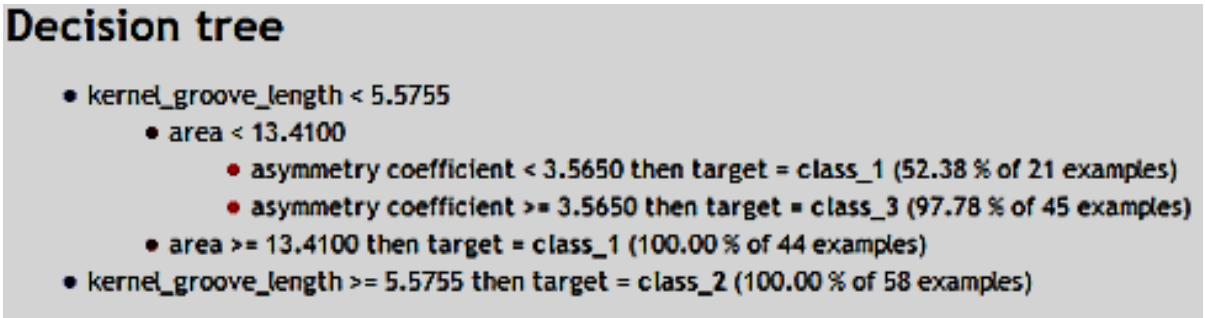

(a) Import trained decision tree.

Fig. 11. Import Seeds data and trained decision tree from Tanagra system.

**Bended DT visualization from imported DT model**. Fig. 11 illustrates importing data and trained DT. Next, Fig. 12. shows on the left the decision tree built using the training data with those data visualized on the DT. On the right it shows the same DTwith the testing data visualized on it. It allows comparing training and testing data side-by-side.= Fig. 12 clearly shows similarity of the distribution of the training and validation data in the DT. They match each of each other quite well. Thus, the accuracy of the DT model on both datasets should be similar. Otherwise, a user can modify DT or reject it.

**Bended DT visualization from scratch**. The next few figures show the process of building DT from scratch either by using existing DT as a prototype or by relying on the expert domain knowledge of the user. First the root node is identified with an attribute, a threshold *T* and an operator at the threshold. In Fig. 13, T=5.5755 and the activated operator is “<”, which means that all x<T will be on the left branch, else x will be on the on the right branch. It immediately shows data split with that threshold.

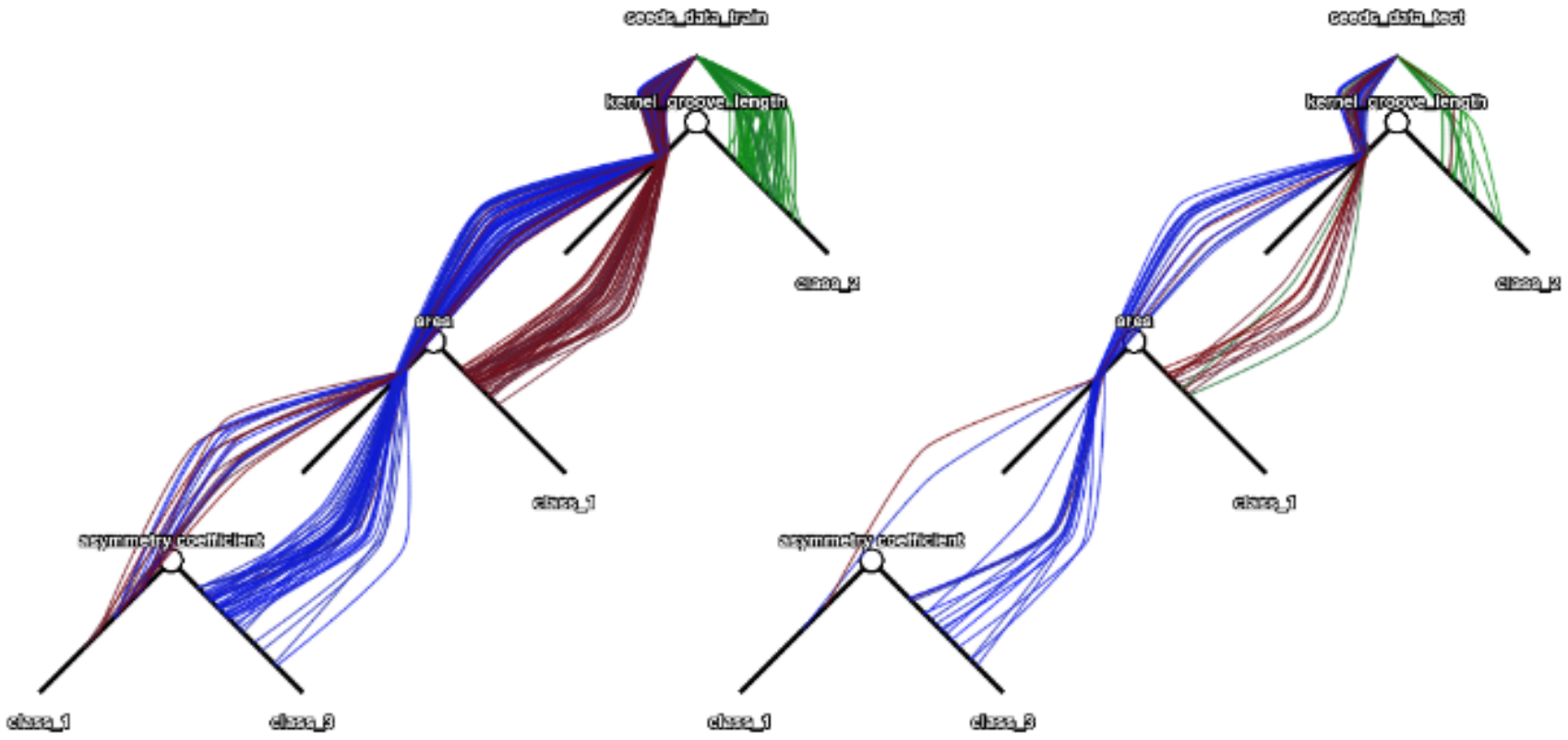

Fig. 12. DT showing both training and testing dataset for seeds dataset for the same tree.

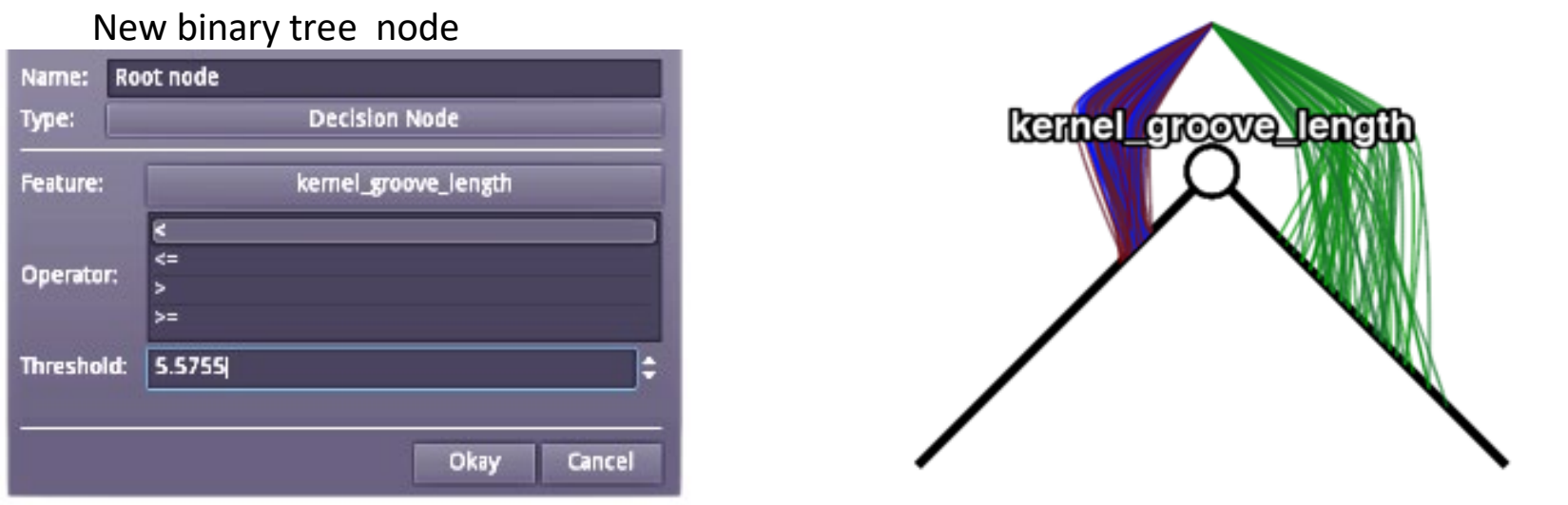

(a) Setting up the root threshold and operator, (b) Decision tree after creating root node.

Fig. 13. Creating root node in the Seeds dataset decision tree.

In Fig. 13, we see that the green cases are well separated from blue and brown cases with that threshold. A user can change a threshold and immediately see how the data split will be changed.

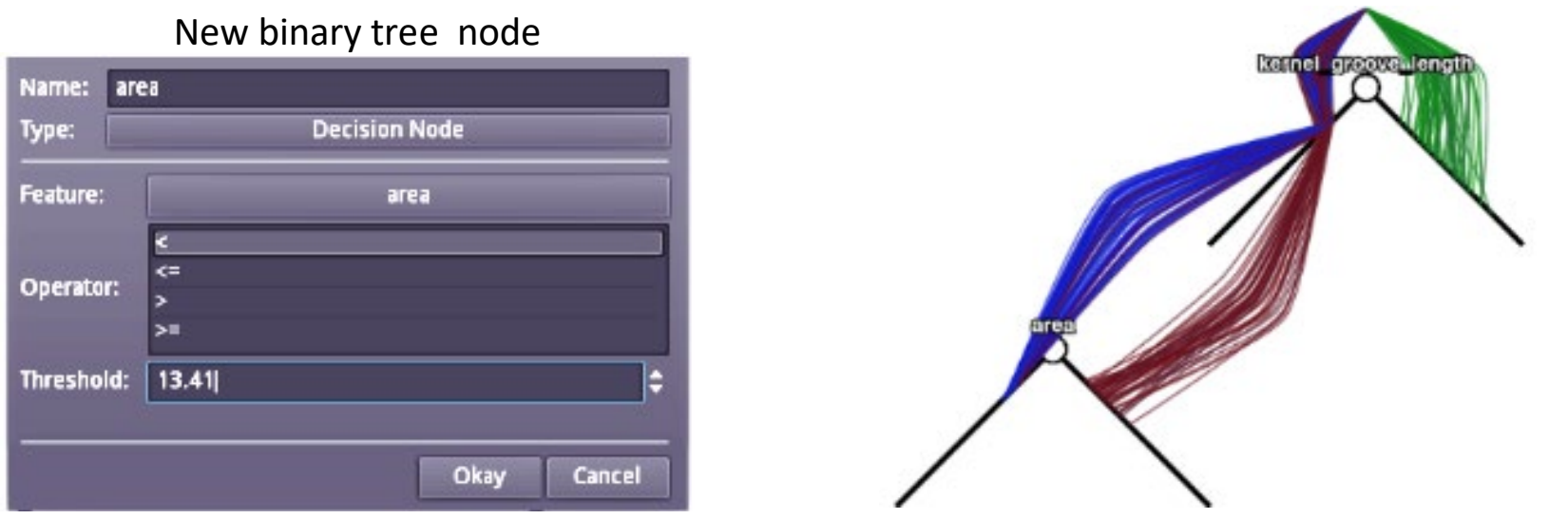

(a) Creating left child branch. (b) After creating left child branch.

Fig. 14. Creating left child branch for Seeds dataset.

Fig. 14 shows the DT growth where the left branch is added. It also shows how the data are split further. So now we can see better that blue and brown cases were not separated well at the first node, in contrast with a new split. Again, changing the threshold at this node allows to see the benefit of different thresholds. The process continues to get a full DT. Figs. 15-16 show the adjustment of the DT thresholds by using a scroll bar.

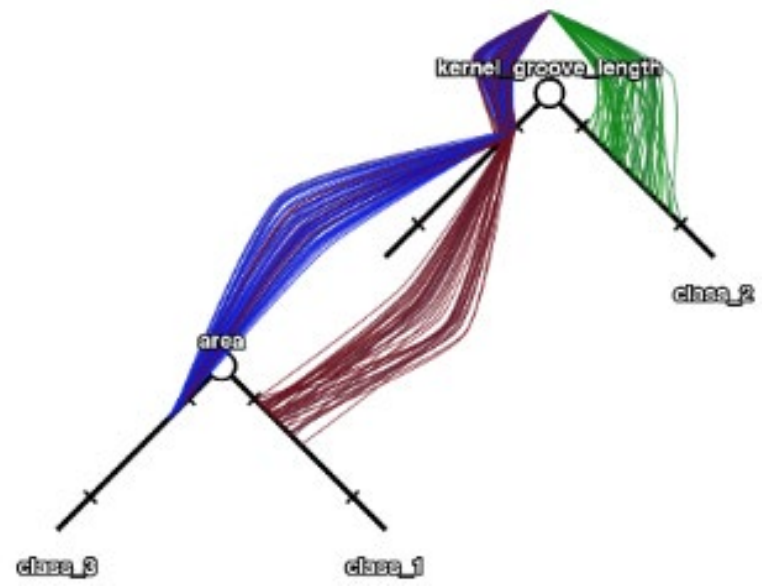

Fig. 15. Base seeds tree with trained threshold.

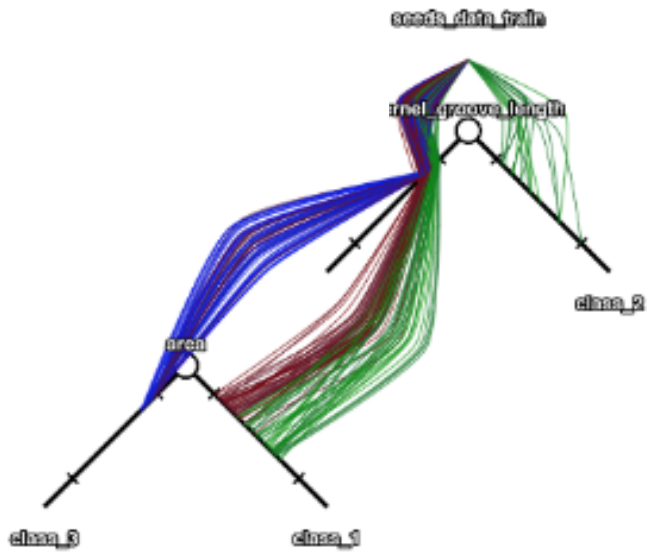

(a) Seeds tree with modified root threshold dragged to the right.

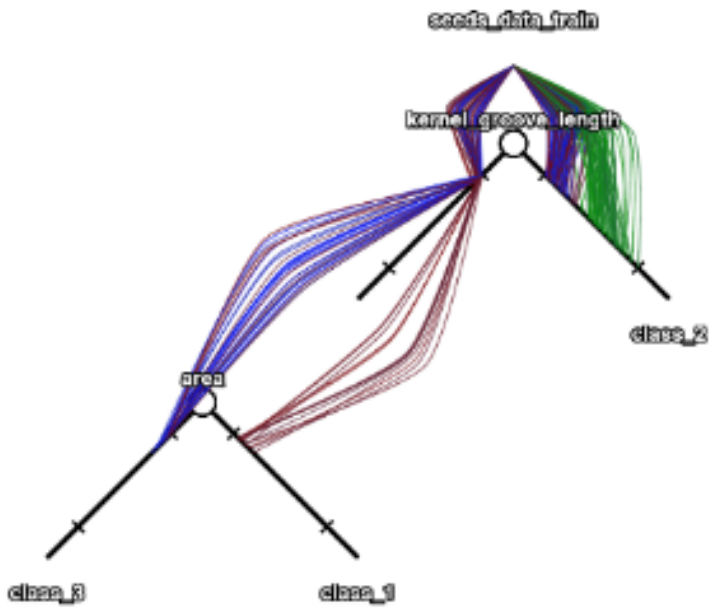

(b) Seeds tree with modified root threshold dragged to the left.

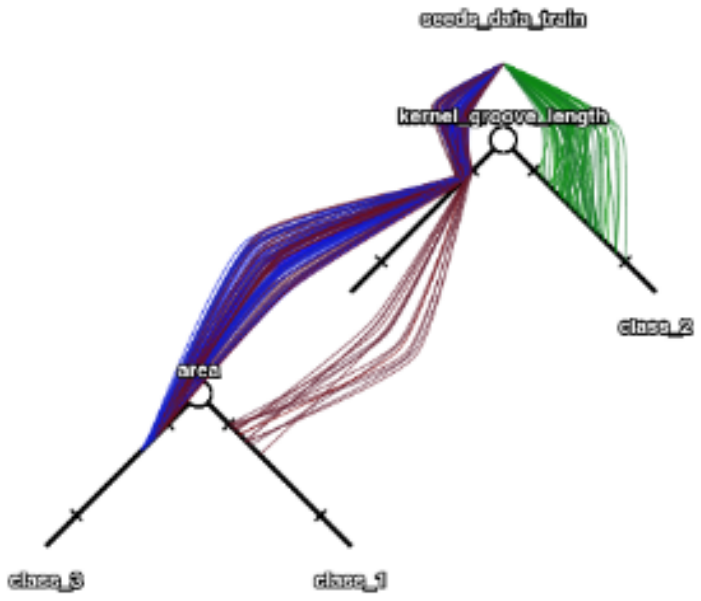

(c) Seeds tree with modified child tree threshold dragged to the right.

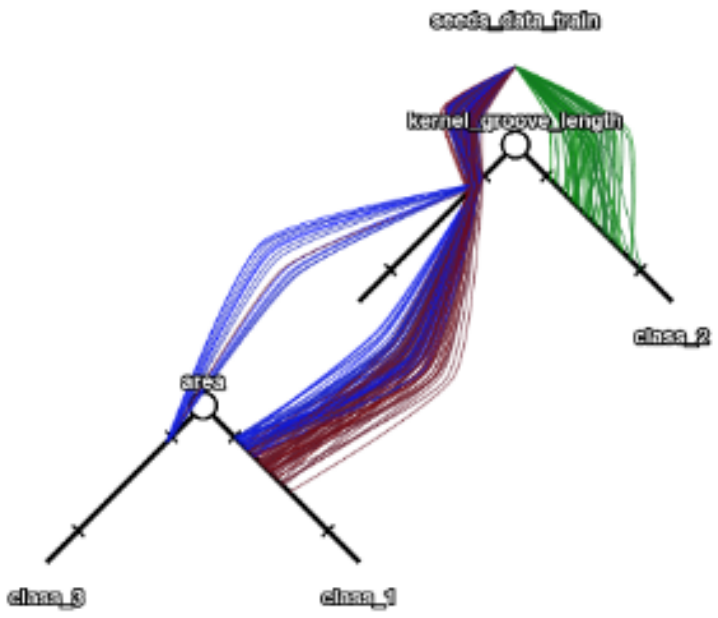

(d) Seeds tree with modified child tree threshold dragged to the left.

Fig.16. Interactive decision tree threshold modification for Seeds data.

### 3.2. Case study: Wisconsin Breast cancer dataset

For the Wisconsin Breast cancer dataset, the experiment shows the abilities of SPC-DT to visualize the DT and to adjust the threshold interactively to build a better DT, which will not misclassify cancer cases as benign.

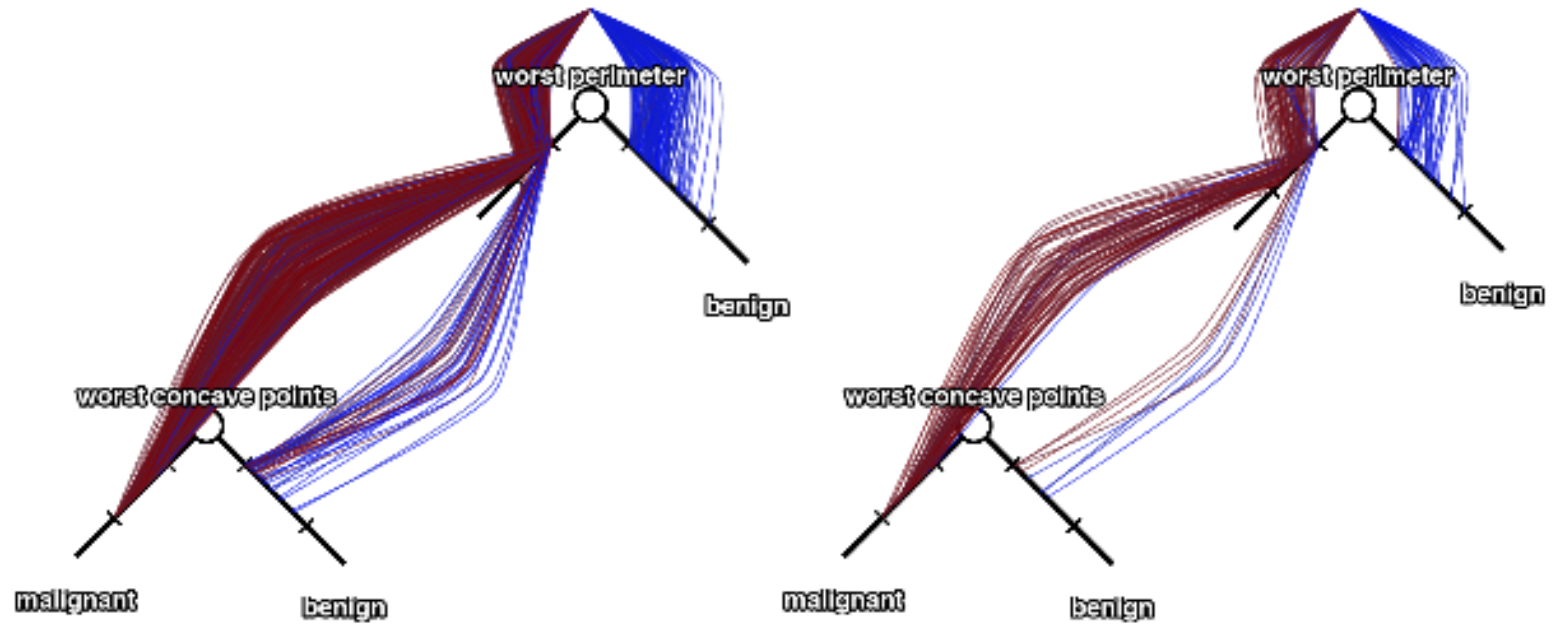

Fig. 17. Trained DT with cancer training data (left) and test data (right) visualized on the DTs.

While it allowed it improvement in the number of false negatives, it's not enough to remove all false negatives in test data. Fig. 17 shows a trained DT with the thresholds identified by the traditional automatic DT training algorithm with the names of attributes listed above the nodes. Fig. 18 depicts the quality of this DT on training and test data, with 19 malignant cases misclassified as benign cases in the training data and 6 malignant cases misclassified as benign cases on the test data. Both failure cases are visible in Fig. 17.

| Training data | | | Test data | | |
|---|---|---|---|---|---|
| Accuracy | | 93.85 | Accuracy | | 92.11 |
| Precision | | 94.44 | Precision | | 94.00 |
| Recall | | 88.95 | Recall | | 88.68 |
| F1 score | | 0.9162 | F1 score | | 0.9126 |
| Classified | | 455 | Classified | | 114 |
| Confusion Matrix | | | Confusion Matrix | | |
| | benign | malignant | | benign | malignant |
| benign | 153 | 9 | benign | 47 | 3 |
| malignant | 19 | 274 | malignant | 6 | 58 |

Fig.18. Confusion matrices for training and testing data for DT from Fig. 17.

We want to decrease these false negatives by building a better DT interactively in the BC-DT software tool. The first attempt was conducted by interactively adjusting the threshold at the root of the left DT that shows training data. The results are presented in Figs. 19-20.

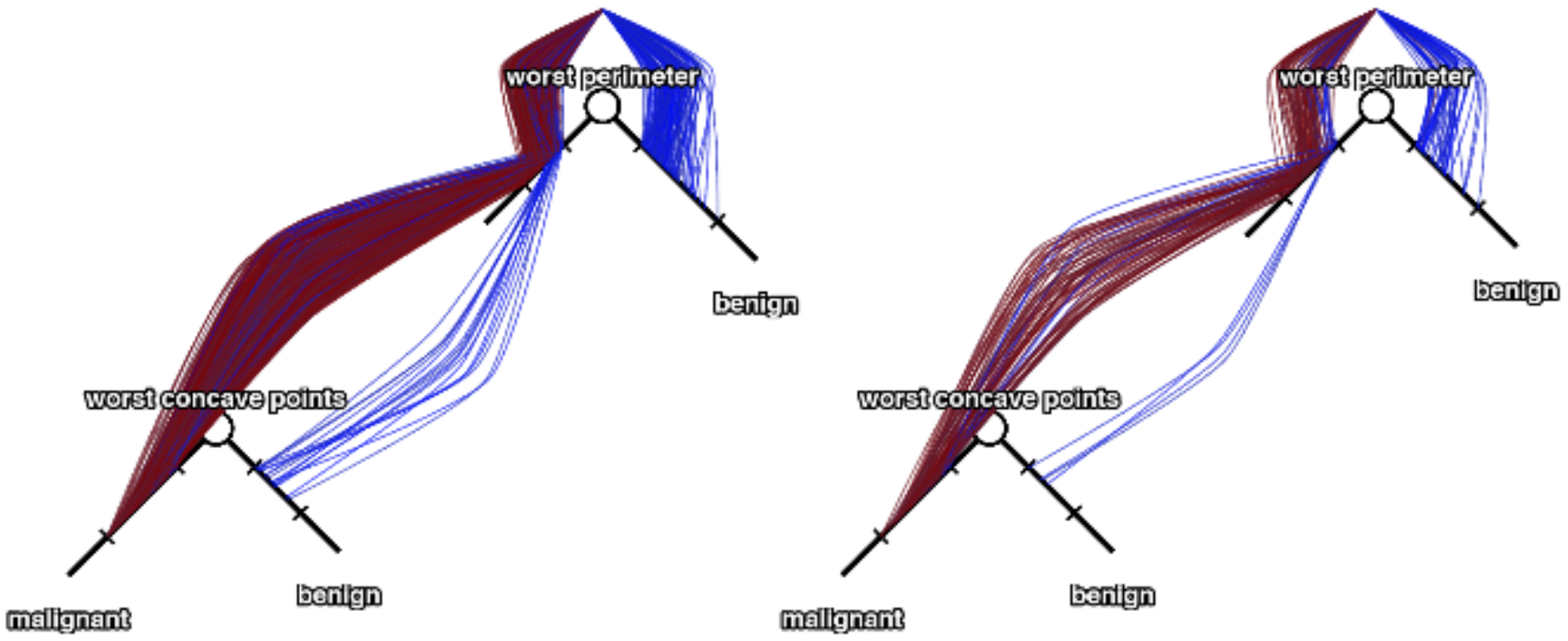

Fig.19. DT with modified root threshold showing improvement to false negatives for cancer data.

| Training data | | | Test data | | |
|---|---|---|---|---|---|
| Accuracy | | 92.97 | Accuracy | | 92.98 |
| Precision | | 80.25 | Precision | | 86.00 |
| Recall | | 100.00 | Recall | | 97.72 |
| F1 score | | 0.8904 | F1 score | | 0.9149 |
| Classified | | 455 | Classified | | 114 |
| Confusion Matrix | | | Confusion Matrix | | |
| | benign | malignant | | benign | malignant |
| benign | 130 | 32 | benign | 43 | 7 |
| malignant | 0 | 293 | malignant | 1 | 63 |

Fig. 20. Confusion matrices for the decision tree from Fig. 19.

The next attempt is presented in Figs. 21 and 22. The DT with modified root and subtree thresholds, shows elimination of false negatives on training data more improvement to training, but there is still one false negative for the testing dataset. The DT with root threshold modified a second time to further improve testing dataset false negatives, which now both datasets have zero false negatives.

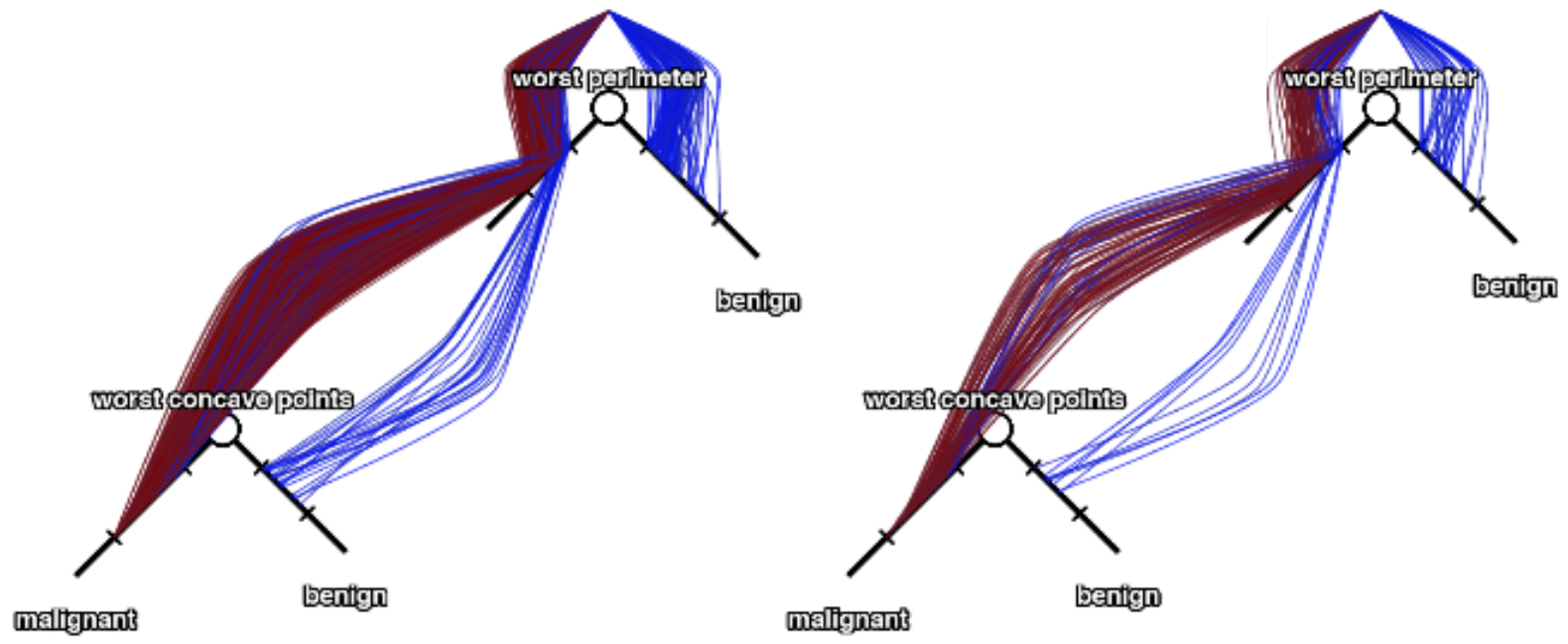

Fig, 21.DT with the fully eliminated false negatives for breast cancer data.

| Training data | | | Test data | | |
|---|---|---|---|---|---|
| Accuracy | | 91.84 | Accuracy | | 91.23 |
| Precision | | 76.54 | Precision | | 80.00 |
| Recall | | 100.00 | Recall | | 100.72 |
| F1 score | | 0.8671 | F1 score | | 0.8888 |
| Classified | | 455 | Classified | | 114 |
| Confusion Matrix | | | Confusion Matrix | | |
| | benign | malignant | | benign | malignant |
| benign | 124 | 38 | benign | 40 | 10 |
| malignant | 0 | 293 | malignant | 0 | 64 |

Fig. 22. Confusion matrices for the decision tree from Fig. 21.

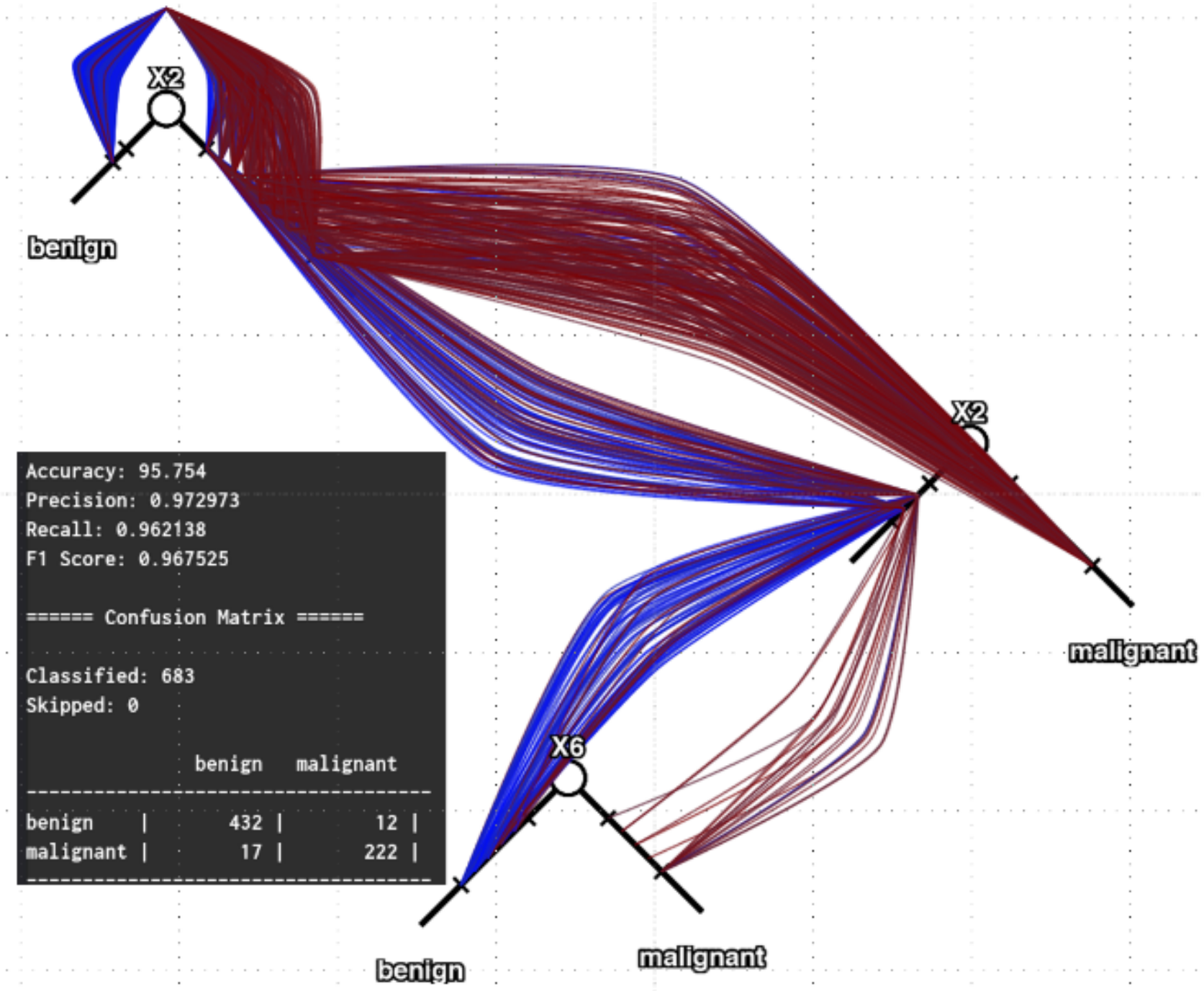

Fig. 23. DT trained on 683 Wisconsin Breast Cancer cases with confusion matrix

Fig. 23 presents a DT trained on all 683 Wisconsin Breast Cancer cases. It contains edges bended in the threshold points with unequal segments allowing to see actual the distribution of attribute values. Inequality is especially visible for attribute x2. Also, it is visible that thresholds are stable because there are no points next to the thresholds.

## 4. SPC-DT Case Studies

This section presents visualization experiments with SPC-DT on datasets from UCI ML Repository [1].

### 4.1. Cases Study: Wisconsin Breast Cancer

With the Wisconsin breast cancer dataset, we continue our investigation of the SPC-DT approach in this section. Our aim is to reduce the overlap of the lines to improve the visualization that was previously presented in Fig. 10 This will make it simpler to follow specific cases, particularly when there are several gray areas in the coordinate pairs. Using the interactive features of the SPC-DT software application, Fig. 24 displays the same DT as in Fig. 10 with the individual coordinate pairs moved and rearranged to improve the SPC-DT visualization.

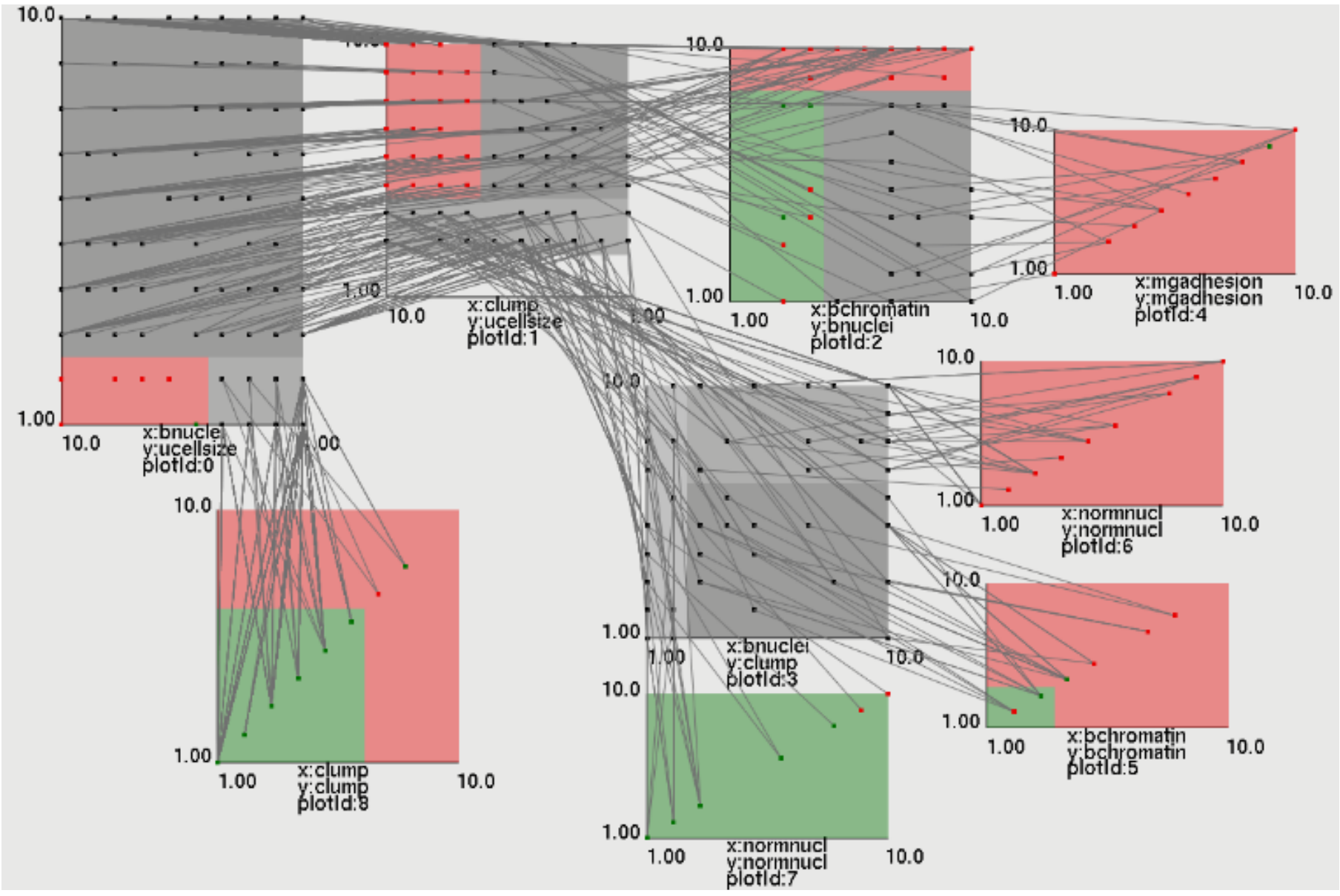

. Fig. 24. WBC data in SPC-DT with rearrangement and relocation.

This reorganization and placement of coordinate pair plots greatly increased the DT's analysis clarity. For this, care has been taken to minimize line overlap. Yet there's still room for improving SPC-DT visualization clarity. In several coordinate pair graphs, lines overlap significantly, and many cases cross areas to which they do not belong.

Fig. 25 shows the same data as Fig. 24 and Fig. 10, but with the **condensation** of points in gray zones, separated by each case's class. This allows for a significant increase of clarity, especially in coordinate pair plots that have a high number of cases passing through them, like the first plot of the left in Fig. 25, as well as plots that have a high number of gray zones like in other plots in this figure. Yet, the amount of data that may be losslessly shown is inevitably reduced when condensation is used. Condensing all points in a 2D space to a single point obscures the fact that the classification can become more uncertain the closer a point approaches the boundaries of classification zones.

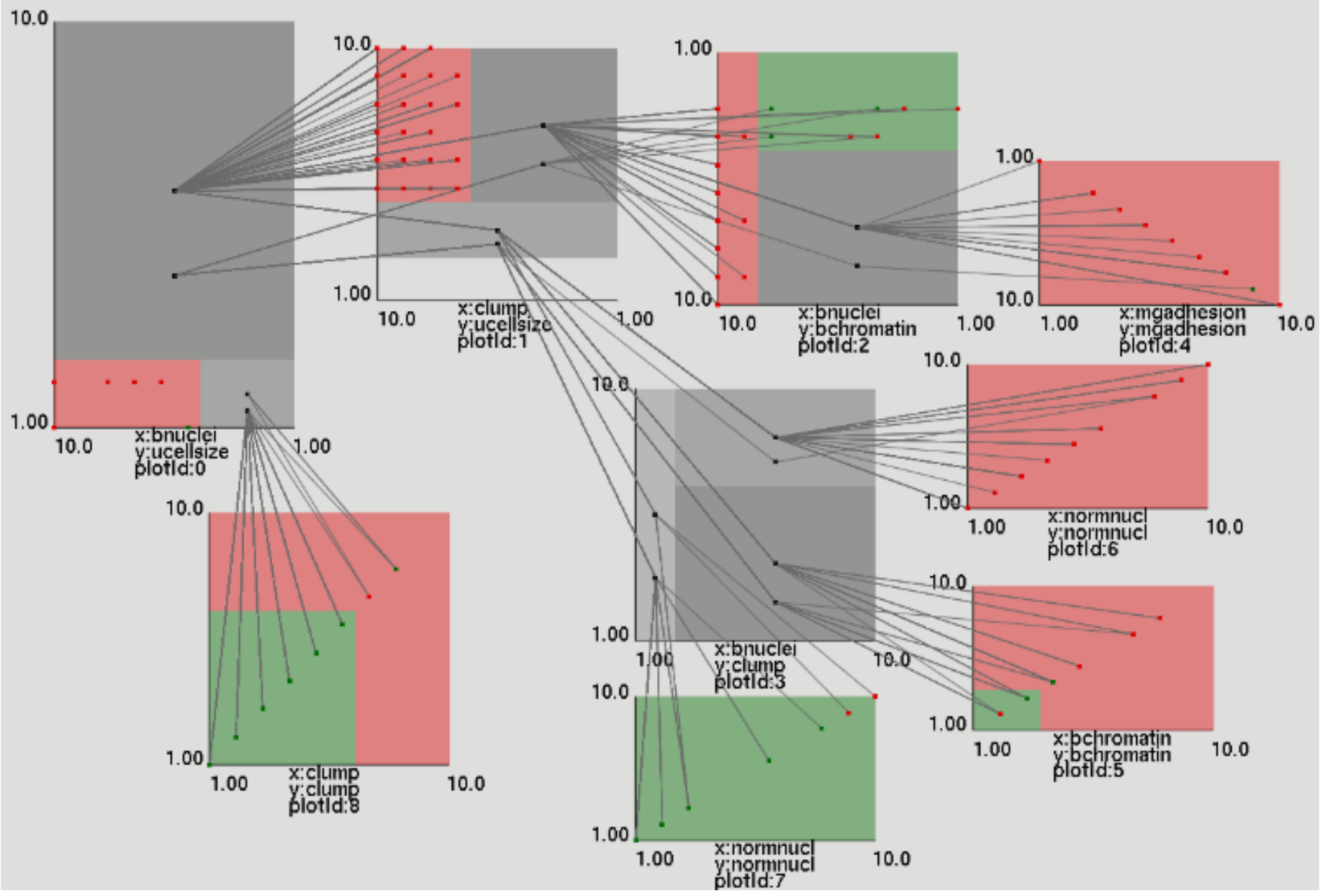

Fig. 25. WBC data in SPC-DT with rearrangement, relocation, and condensation.

### 4.2. Case Study: Iris Dataset

There are 150 instances of each class in this dataset. Four attributes are provided for each example. The decision tree used to visualize the Iris data in SPC-DT is shown in Fig. 26. Its performance is shown in Table 3. The DT using the SPC-DT approach is visualized in Fig. 27 with all the Iris data in the initial places for each coordinate pair.

- petal-length < 2.4500 then class = **Iris-setosa** (100.00 % of 50 examples)
- petal-length >= 2.4500
  - petal-width < 1.7500
    - petal-length < 4.9500
      - sepal-width < 2.5500
        - sepal-width < 2.4500 then class = **Iris-versicolor** (100.00 % of 9 examples)
        - sepal-width >= 2.4500 then class = **Iris-versicolor** (80.00 % of 5 examples)
      - sepal-width >= 2.5500 then class = **Iris-versicolor** (100.00 % of 34 examples)
    - petal-length >= 4.9500 then class = **Iris-virginica** (66.67 % of 6 examples)
  - petal-width >= 1.7500
    - sepal-length < 5.9500 then class = **Iris-virginica** (85.71 % of 7 examples)
    - sepal-length >= 5.9500 then class = **Iris-virginica** (100.00 % of 39 examples)

Fig. 26. ID3 Decision Tree for Iris data.

The gray areas in Fig. 27 represent regions within each paired coordinate space where it is impossible to define a case's class at the current DT level and where it must be followed to the next DT level to be evaluated further.

The green areas denote cases that the DT found to be "Virginica," the blue areas denote cases that the DT classed as "Versicolor," and the red areas denote situations that the DT identified as "Setosa." As a class has already been established, these cases end at the current level. Due to the extensive line overlap, Fig. 27 does not clearly depict patterns. Although it is simpler to follow individual examples than in Fig. 24, line congestion still exists.

TABLE 3. PERFORMANCE OF TREE IN FIG. 26

| Error rate | | | 0.0267 | | | |
|---|---|---|---|---|---|---|
| Values prediction | | | Confusion matrix | | | |
| Value | Recall | 1-Precision | | setosa | versicolor | virginica |
| setosa | 1.00 | 0.0000 | setosa | 50 | 0 | 0 |
| versicolor | 0.9461 | 0.0208 | versicolor | 0 | 47 | 3 |
| virginica | 0.9800 | 0.0577 | virginica | 0 | 1 | 49 |

The repositioning of coordinate pair plots results in a clearer image, as seen in Fig. 28.

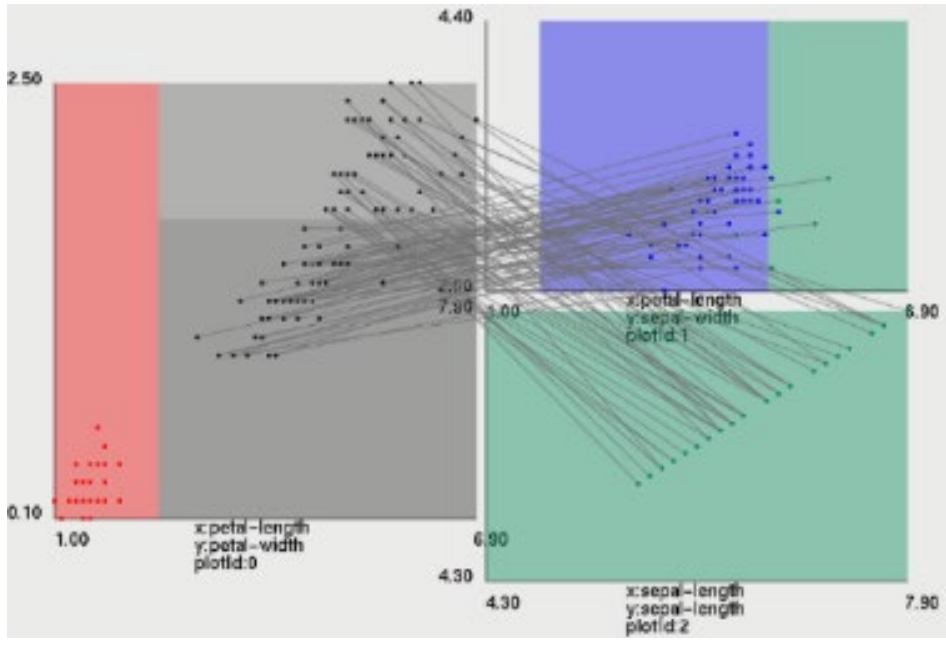

Fig. 27. Iris in default SPC-DT.

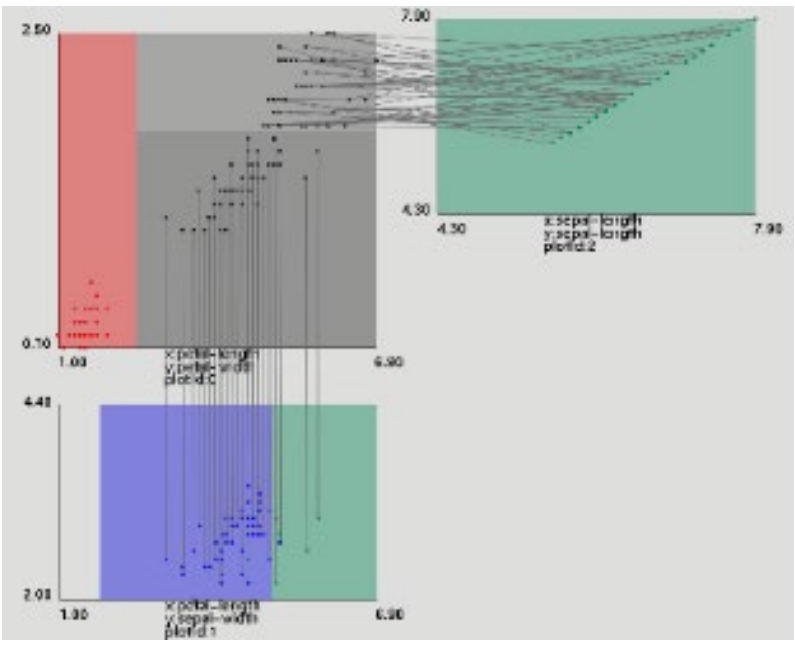

Fig. 28. Iris data in SPC-DT relocated.

This method of moving them lessens the number of line crossings and removes the paths that go through classification zones that don't correspond to the target class. Line condensation with Iris data is shown in Fig. 29. The DT incorrectly classified the red points in plots 1 and 2. The plots were moved and adjusted to take advantage of the gray point condensation. It improved visibility of instances that were incorrectly classified and their path to their final class in the DT.

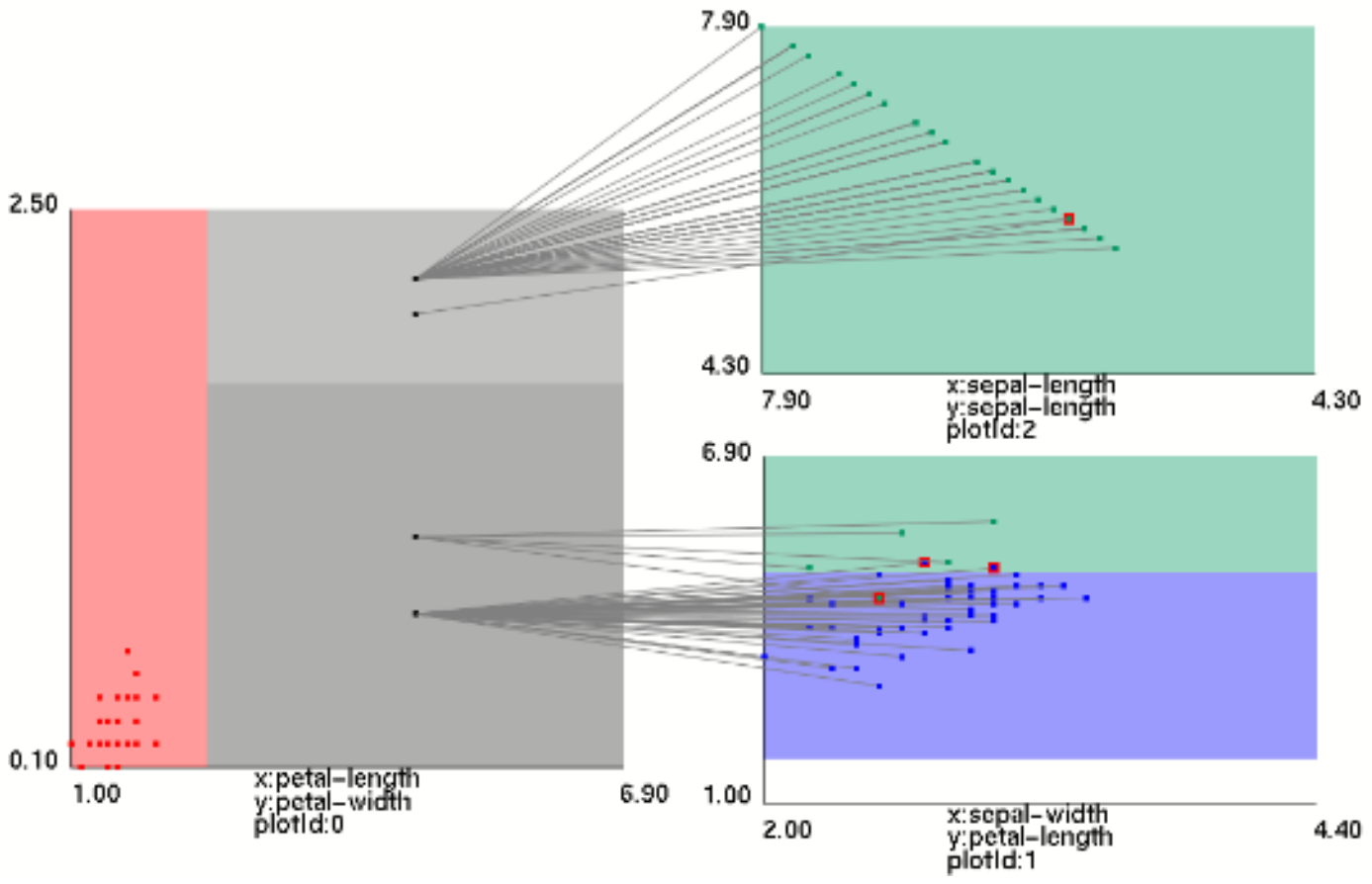

. Fig. 29. Iris in SPC-DT relocated, rearranged, and condensed.

### 4.3. Case study: Wine Dataset

The wine dataset holds 178 cases of 3 classes. Each case is presented by 13 attributes. Fig. 30 and Table 4 describe the DT used.

- Flavanoids < 1.5750
  - Color-intensity < 3.8250 then class = **class_2** (100.00 % of 13 examples)
  - Color-intensity >= 3.8250
    - Toal-Phenols < 2.0100 then class = **class_3** (100.00 % of 42 examples)
    - Toal-Phenols >= 2.0100 then class = **class_3** (85.71 % of 7 examples)
- Flavanoids >= 1.5750
  - Proline < 724.5000
    - Malic-Acid < 3.9250 then class = **class_2** (100.00 % of 49 examples)
    - Malic-Acid >= 3.9250 then class = class_2 (80.00 % of 5 examples)
  - Proline >= 724.5000
    - Alcohol < 13.0200 then class = **class_2** (66.67 % of 6 examples)
    - Alcohol >= 13.0200 then class = **class_1** (100.00 % of 56 examples)

Fig. 30. ID3 Decision tree for Wine data.

TABLE 4. PERFORMANCE OF TREE IN FIG. 30.

| Error rate | | | 0.0225 | | | |
|---|---|---|---|---|---|---|
| Values prediction | | | Confusion matrix | | | |
| Value | Recall | 1-Precision | | Class_1 | Class_2 | Class_3 |
| Class_1 | 0.9492 | 0.0000 | Class_1 | 56 | 3 | 0 |
| Class_2 | 0.9859 | 0.0411 | Class_2 | 0 | 70 | 1 |
| Class_3 | 1.000 | 0.0204 | Class_3 | 0 | 0 | 48 |

All the data in SPC-DT are displayed in their default places in Fig. 31.

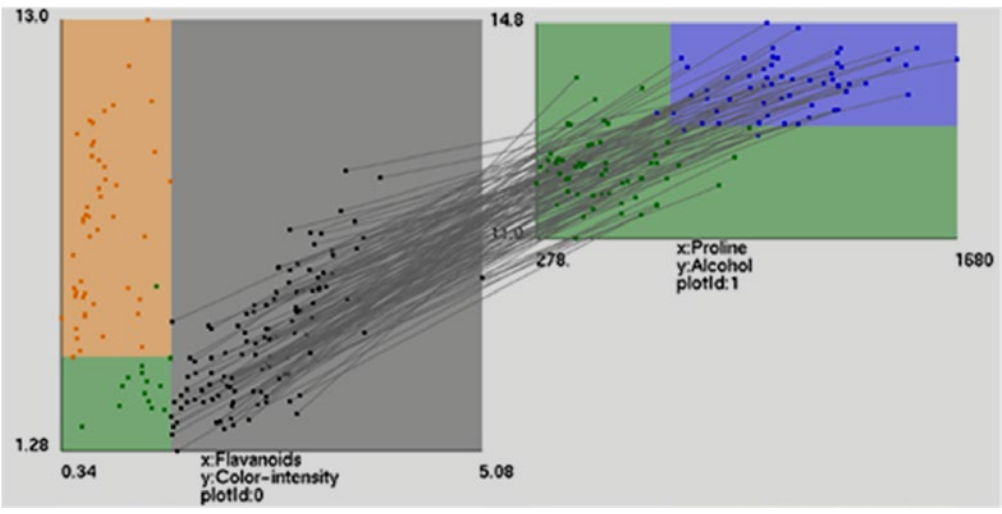

Fig. 31. Default Wine dataset in SPC-DT.

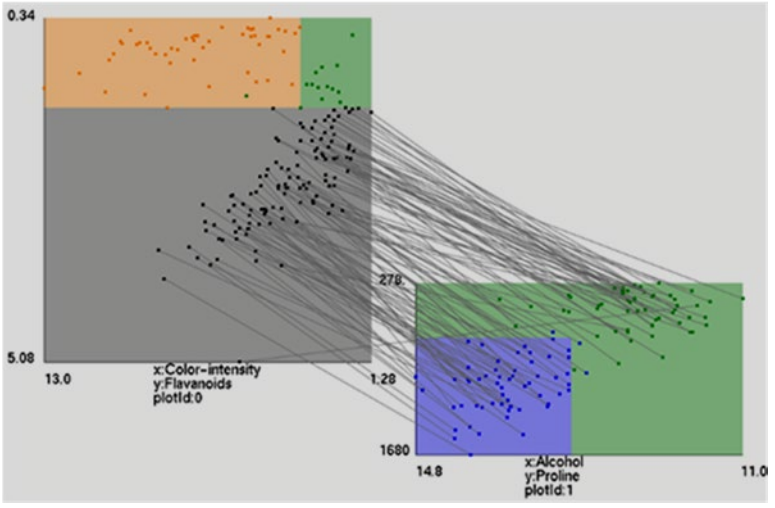

Fig. 32. Rearranged Wine dataset in SPC-DT.

The gray region denotes areas in each paired coordinate space where it is impossible to determine a case's class at the current DT level and where it must be followed to the following DT level to be evaluated further.

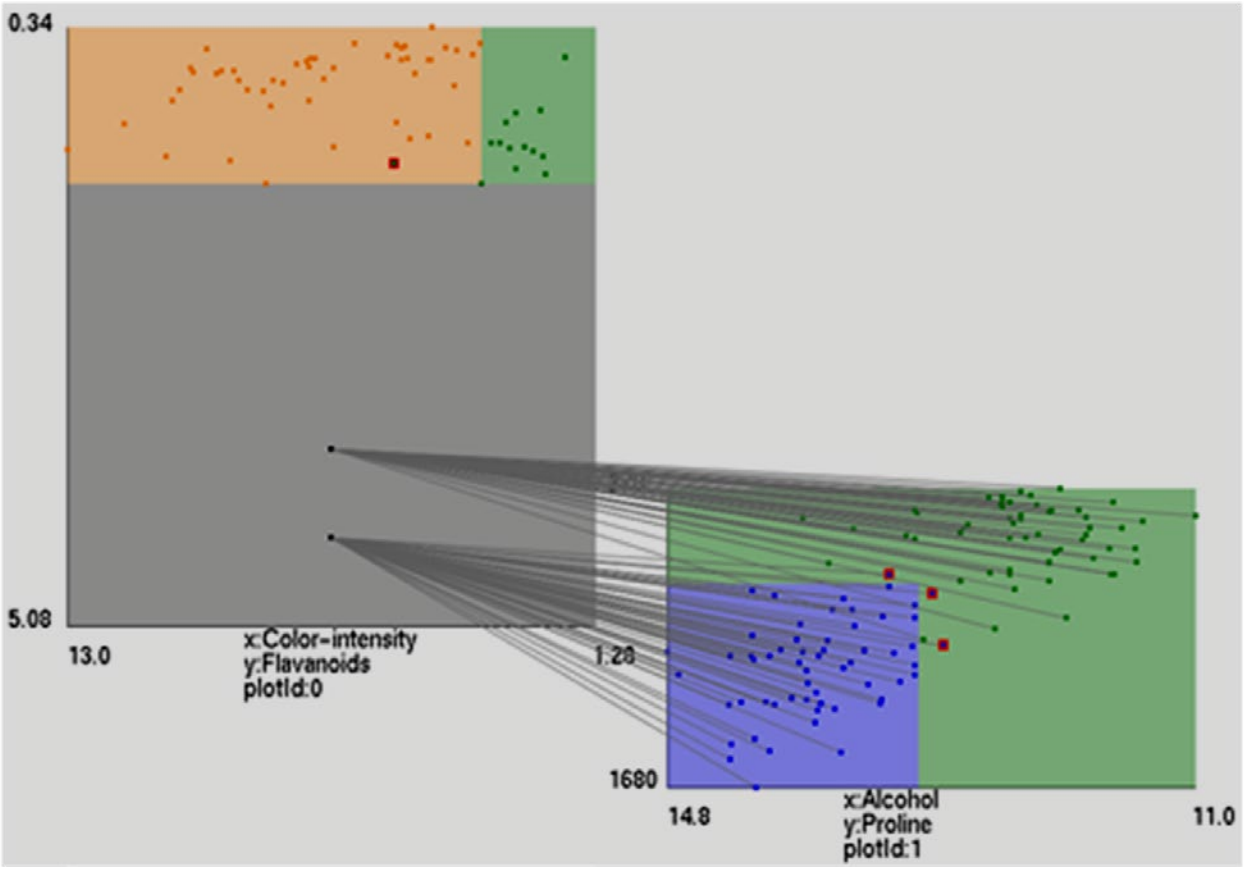

Fig. 33. Wine dataset in SPC-DT with relocation, rearrangement, and condensation.

After moving the plots and reordering their axes, Fig. 32 exhibits a minor gain in clarity. As a result, fewer lines cross through classification zones that do not correspond to the class of their destination, but fewer overlapping lines are created. The orange region denotes cases that the DT has classified as "class 1", the green area denotes cases that the DT has classified as "class 2", and the blue area denotes cases that the DT has classified as "class 3." As a class has already been established, these cases end at the current level. The default visualization of the Wine dataset exhibits significant line overlap across coordinate pair plots, just as the other case studies. By combining relocation, rearrangement, and condensation, a method similar to that employed on the Iris and WBC datasets can be utilized to reduce this overlap.

The Wine data in SPC-DT are displayed in Fig.33. The gray area in the plot with the caption "plotId: 0" becomes significantly clearer after applying condensation since there is less line overlap. Misclassifications are highlighted in Fig. 33 as the points with the red frames, which adds more clarity. This can make it easier to comprehend how well the decision tree is performing while being visualized.

### 4.4. Case Study: WBC Dataset with split Training and Validation Data

The benefits of using the SPC-DT approach to visualize datasets that are divided into training and validation data are demonstrated in this section. The split makes it possible to observe and examine relationships between training and validation data that do not pertain to training data. Separating data can improve the DT model's confidence or, alternatively, can result in updating and redesigning the decision tree, which speeds up the DT modeling process. Table 5 and Fig. 34 describe the decision tree.

Data visualizations for training and validation are shown in Figs. 35 and 36, respectively. Darker classification color shades are utilized in these representations to indicate more examples that the DT rule covers. More cases are indicated by darker colors. The white borders around some cases are there to make them more visible.

- ucellsize < 2.5000
  - sepics < 2.5000 then class = **begnin** (99.43 % of 349 examples)
  - sepics >= 2.5000 then class = **begnin** (77.42 % of 31 examples)
- ucellsize >= 2.5000
  - ucellsize < 4.5000
    - bnuclei < 3.5000 then class = **begnin** (78.38 % of 37 examples)
    - bnuclei >= 3.5000 then class = **malignant** (85.11 % of 47 examples)
  - ucellsize >= 4.5000
    - clump < 6.5000
      - bnuclei < 8.5000 then class = **malignant** (80.00 % of 25 examples)
      - bnuclei >= 8.5000 then class = **malignant** (100.00 % of 41 examples)
    - clump >= 6.5000 then class = **malignant** (100.00 % of 99 examples)

Fig. 34. ID3 Decision Tree of 90/10 split WBC Data.

To aid in the visualization of misclassified cases, overlapping classifications have been spread out. Tightly packed cases that appear in an upper left to lower right diagonal line all contain the same value, but they have been spread to better visualize misclassified cases that may be hard to see otherwise. A confusion matrix has been included in Fig. 35 because the confusion matrix in Fig. 34 is for the 90% training dataset.

TABLE 5. PERFORMANCE OF TREE IN FIG. 34

| Error rate | | | 0.046 | | | |
|---|---|---|---|---|---|---|
| Values prediction | | | Confusion matrix | | | |
| Value | Recall | 1-Precision | | benign | malignant | Sum |
| benign | 0.9709 | 0.0408 | benign | 400 | 12 | 412 |
| malignant | 0.9217 | 0.0566 | malignant | 17 | 200 | 217 |
| | | | Sum | 417 | 212 | 629 |

Interestingly, the decision tree's structure is different from the DTs for WBC presented above. This is because earlier DTs were constructed using 100% of the WBC dataset, but the DT in Figs. 35 and 36 were constructed using just 90% of the WBC dataset. This variation shows how the DT models are unstable. Although though DT instability is a well-known feature, it can be challenging to distinguish the differences between the traditional DT visualizations preseted above and the SPC-DT representation.

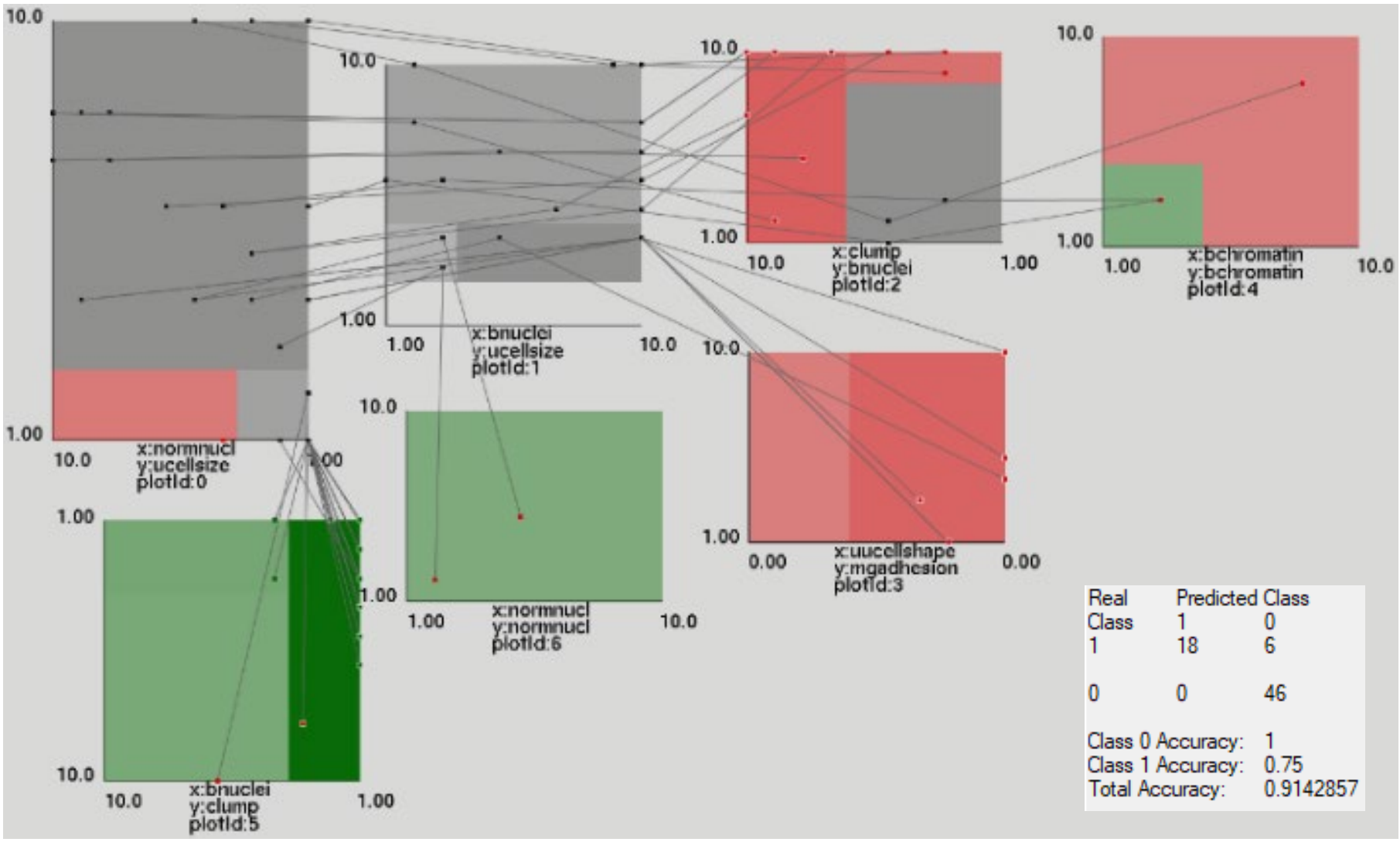

Fig. 35.Testing data in 90/10 split DT in SPC-DT.

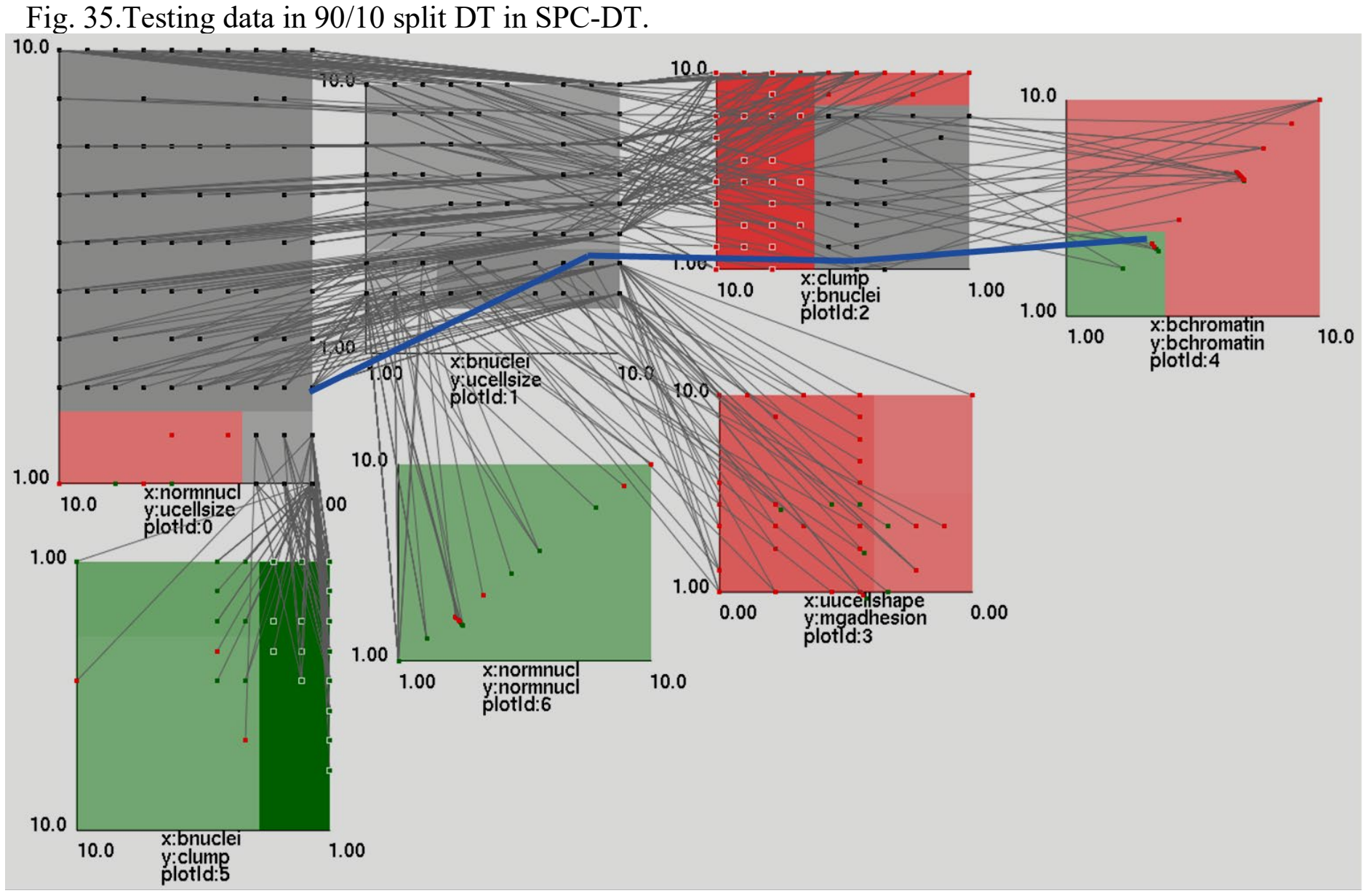

. Fig. 36. Training data in 90/10 split DT in SPC-DT.

The distribution of the cases in the validation data is evidently different from that in the training data. Significant areas that were covered by the training data are not covered by it. As a result, the accuracy of the validation data cannot accurately represent the accuracy of the training data, and the validation data may not accurately represent the training data. Figs. 35 and 36, which have different accuracies: 91.43% on validation data and 95.39% on training data, show this disparity.

The blue case, which begins in the gray area in the left plot very near the green area, is likewise depicted in Fig. 36. It continues to the second plot's gray region until coming to an end in the bottom plot's red area, which is once more quite near the green area. This classification is unreliable, as demonstrated by the case's visual analysis. The classification of this case will alter with a small adjustment to the borderlines. This shows the advantages of visual analysis and the SPC-DT visualization strategy.

The ability to display the actual cases and their 2-D distributions in the $(X_i,X_j)$ coordinates given in the adjacent nodes of the DT is one of the SPC-DT method's key features. It enables for the interactive improvement of the DT model's accuracy through the tracing of incorrectly classified cases. For instance, a user can interactively alter the thresholds in these nodes in the SPC-DT visualization to change the classification of misclassified cases that are close to the threshold and see how it affects the classification of other training and validation cases.

A user can interactively tune the DT using the SPC-DT approach. While certain DT visualizations, as [5], permit interactive optimization using 1-D distributions, SPC-DT permits it more effectively in 2-D distributions. Moreover, SPC-DT offers a more compact visualization of fewer tracing points because each tracing point in the SPC-DT visualization represents two attributes rather than just one as in the traditional DT visualization.

# 5. Discussion and generalization

This paper proposed a new DT visualization method for machine learning. The features and advantages of this method include showing relations between multiple pairs of attributes in each plot, individual cases, the closeness of the cases to the split thresholds in the DT nodes, and the density of the cases in the parts of the n-D space. These features together help in gaining confidence in the DT model interpretability and accuracy for future predictions, avoiding overgeneralization and overfittings.

### 5.1 Benefits of SPC-DT

The advantage of a combination of Shifted Paired Coordinates with a Decision Tree in SPC-DT is in allowing to discover a more efficient classification model than with DT only. Traditional decision tree algorithms search for a best spit in **one attribute** at the time. SPC-DT allows searching for a best spit in a **pair of attributes** at the time. Below we present a way to benefit from this opportunity. Consider a pair of attributes $(X_i,X_j)$, and cases of blue and yellow classes shown as blue and yellow rectangles in Fig. 37. Every point $\mathbf{p}=(x_i, x_j)$ divides $(X_i,X_j)$ space the produces 4 **quadrants** (see Fig. 37a).

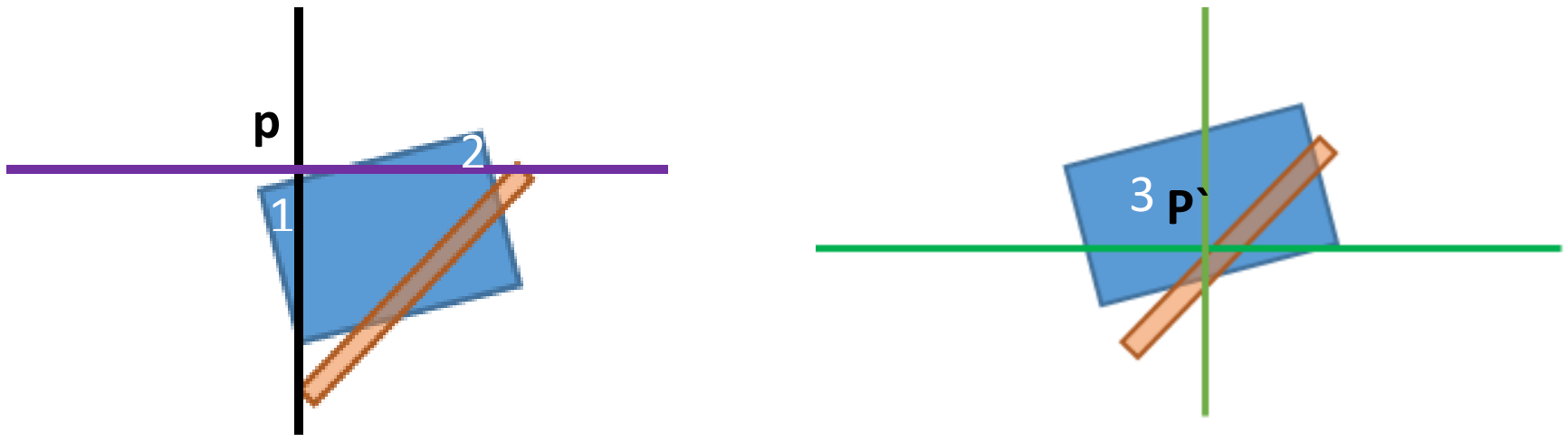

(a) Sequential split: black line then violet line. (b) Simultaneous split by two green lines.
Fig. 37. Sequential (a) vs. simultaneous (b) splits of cases of blue and orange classes.

We want to find point **p** to get the purest separation of blue and yellow, i.e., to maximize the total area all four quadrants where cases belong only to one class by selecting the location of point **p**.

In the **sequential** DT process with n attributes, we order all attributes by purity splits in decreasing order. If split $X_i$ provided the highest purity, then we accept this split and repeat the same purity ordering for each obtained branch. Thus, in the DT split process, the attribute with highest purity in the branch depends on the attribute selected for the parent node. In Fig. 37a the DT algorithm first splits with a black vertical line and then with violet horizontal line generating pure areas 1 and 2. In contrast the SPC-DT algorithm computes purity of 4 quadrants **simultaneously** for each **pair** of attributes. Fig. 37b shows that better split at point **p**`, where area 3 is greater than total areas 1 and 2 in Fig. 37a.

Computationally the search for the point **p**` can be done in multiple ways, including brute force grid search or optimization. Also, this SPC-DT approach can accommodate the different options of defining the purity of the quadrants. including adaptation of the Gini index to quadrants.

The deficiency of the sequential DT process was noticed long time ago and was addressed in the *random forests* algorithms by producing and combining multiple DTs. While it allowed to improve the accuracy of the resulting models it created the problem of *interpretability* for the random forests. In contrast the SPC-DT model is interpretable as a decision tree.

Another deficiency of the decision tree is that each attribute is split only to **two parts** (one-value binary split, $x<T$, $x \geq T$)). A split with more than two parts can be more efficient [14, 28]. In both [14, 28] it is conducted in parallel coordinate, using hyperblocks in [28] and intervals on individual attributes in [14]. Fig. 38 from [14] shows a two-value binary split denoted as split (b).

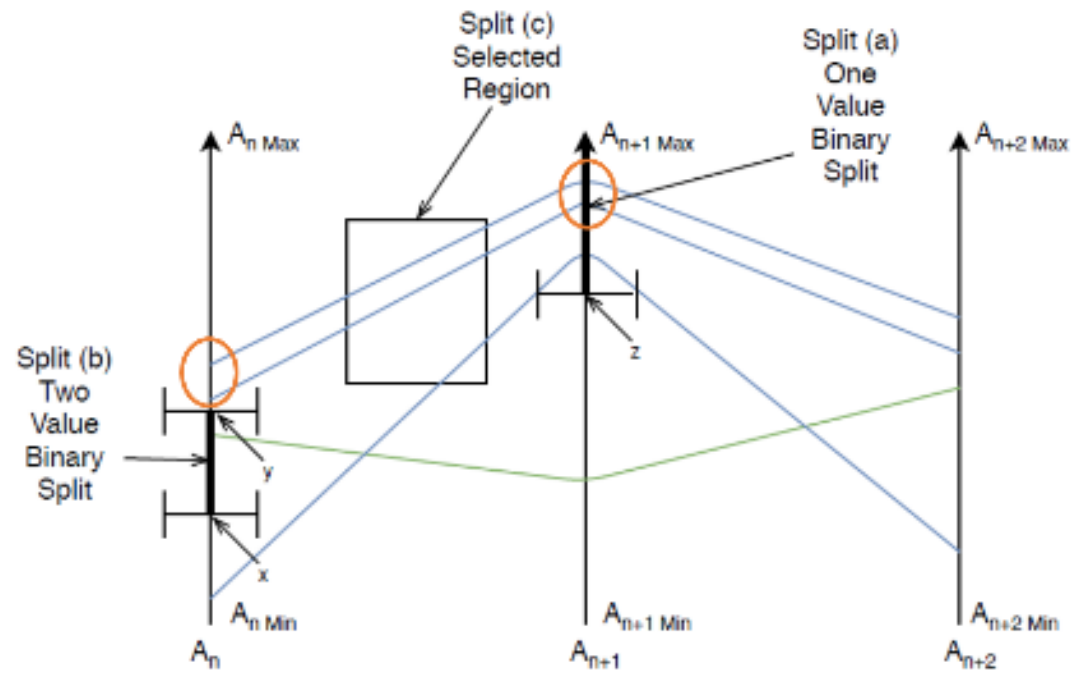

Fig.38. Splitting techniques: (a), DT type simple binary rule x>T for the attribute; (b), a range-based rule, value in [x, y] interval; (c), region-based split for a rule, all cases that pass through the region [14].

This figure also shows another opportunity which is missing in the traditional decision trees. It is selecting the **region** (c) in the parallel coordinates that captures all instances that pass through the region between two adjacent attributes. It captures a relationship between two attributes, while each note of the DT captures a split property of a single attribute. For the region (c), the lines that go through this region can have their points in **multiple intervals** of these adjacent coordinates. This is how it generalizes the decision tree algorithms to richer and more complex structures, models and rules.

In Fig. 38, the red ovals show the intervals in parallel coordinates, which are mathematically equivalent to a rectangle in the Shifted Paired Coordinates. Fig. 39 illustrates this equivalence, where the green quadrilateral in Fig. 39a represents green intervals in $x_1$ and $x_2$ parallel coordinates, respectively.

This green quadrilateral corresponds to the green rectangle in Fig. 39b where $x_1$ and $x_2$ form the orthogonal Cartesian Coordinates, which is a first pair of coordinates in SPC. Similarly, the red quadrilateral in Fig 39a corresponds to a red rectangle in Fig. 39b. Also Fig. 39 illustrates how the black rectangle in Fig. 39a corresponds to the overlap area of the green and red rectangles in Fig. 39b. This single simple black rectangle R represents a more complex rule than captured by a decision tree. For example, a simple visual rule with a rectangle:

$$\text{If } \mathbf{x} \in R \text{ then } \mathbf{x} \in \text{Class1}$$

is equivalent to the rule

$$\text{If } (a_{G11} \leq \boldsymbol{x}_1 \leq a_{G12} \ \& \ a_{G21} \leq \boldsymbol{x}_2 \leq a_{G22}) \& (a_{R11} \leq \boldsymbol{x}_1 \leq a_{R12} \ \& \ a_{R21} \leq \boldsymbol{x}_2 \leq a_{R22}) \Rightarrow \mathbf{x} \in \text{Class1},$$

where $a_{Gij}$ and $a_{Rij}$ limits of the green and red intervals on attributes $x_1$ and $x_2$.

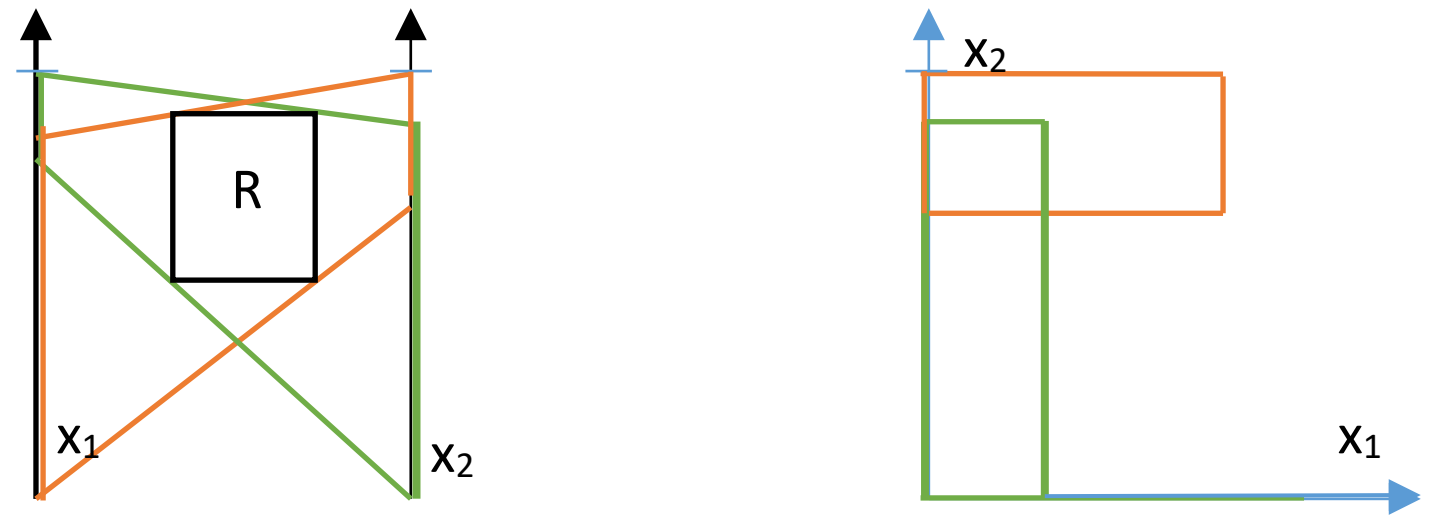

(a) Intervals for the rectangle from Fig. 38 in parallel coordinates.

(b) Red and green rectangles for the black rectangle from (a) in Cartesian Coordinates (first pair of SPC).

Fig.39. Rectangular rule in Parallel Coordinates (a) and in Cartesian Coordinates (first pair of SPC) (b)

The advantage of SPC visualization in Fig. 39b relative to Parallel coordinates is that it requires two times less 2-D points to visualize losslessly each n-D point [3] leading to less occlusion.

The same idea of the rectangular regions to build rules that are more general than DTs was used in [18, 19] with SPC, in [20] with Elliptic Paired Coordinates (EPC) in [20] and with In-Line Coordinates (ILC) in [21]. This approach is expandable to other general line coordinates too.

The advantage of implementing rectangles in SPC is that they are directly visible in the decision tree itself and allow interactively expand and improve the decision tree. Fig. 40 shows this in in SPC-DT. It expands Fig. 29 with forming an impure grey rectangle in the middle of a larger area with mixed cases. This gray rectangle captures cases, which belong to different classes within a larger rectangle. It generalizes a DT by allowing rules like: If **x** ∈ green area then **x** ∈ green class, if **x** ∈ red area then **x** ∈ red class, if **x** ∈ grey area then **x** not classified (refused). This modification of the decision tree allows to avoid misclassification of the cases, which are in the grey rectangle.

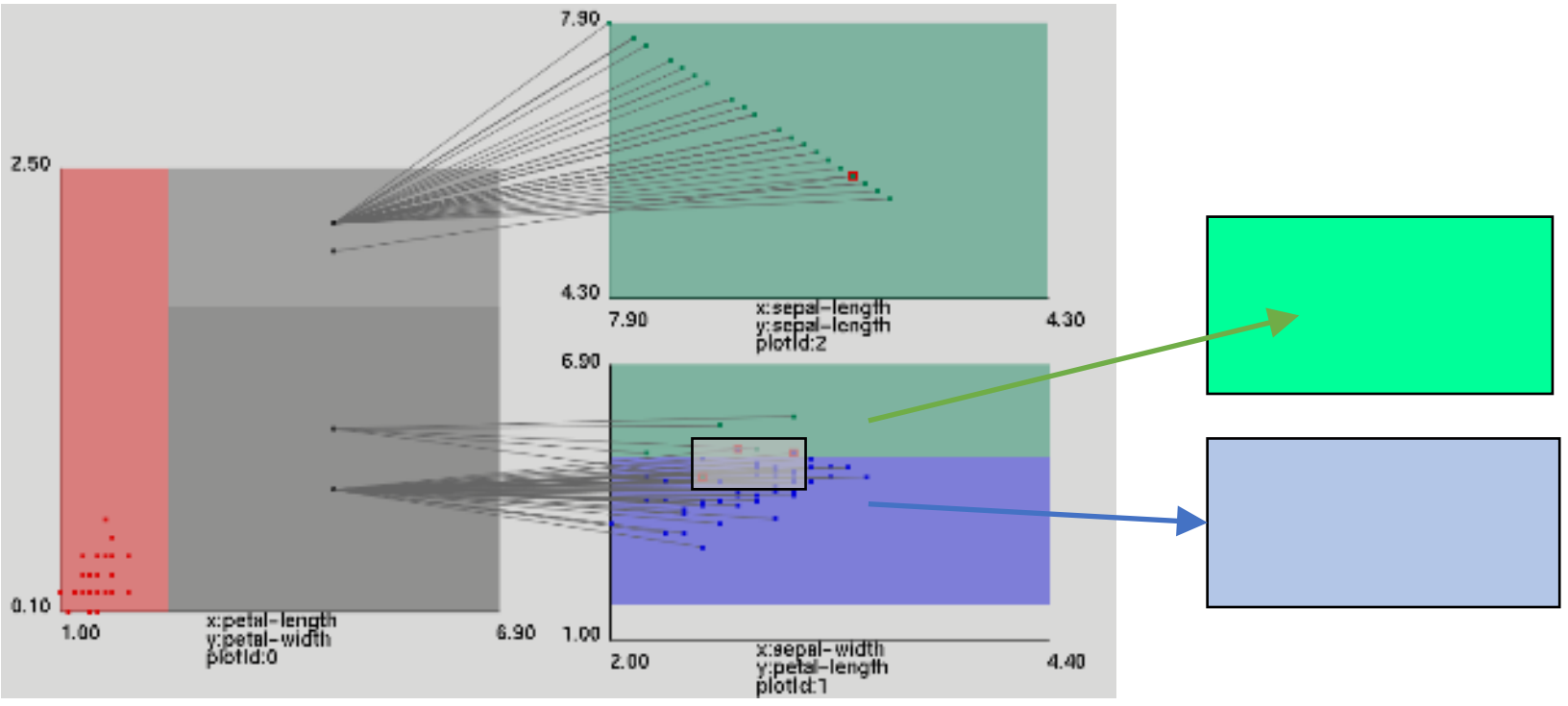

Fig. 40. Expansion of Fig. 29 for impure grey rectangle with black frame in SPC-DT.

It is also can be implemented in Parallel Coordinates as in (c) in Fig. 38, or in PC-DT (see Fig. 5) where DT is combined with Parallel coordinates directly. It was shown in [14] that using parallel coordinates allows to build a better model than traditional decision trees due to generalization of the decision trees.

### 5.2. Other related work

Typically, alternative DT visualizations in machine learning show the number of cases in each node and/or one-dimensional (marginal) distribution of the cases. Fig. 2 in section 1 illustrates such alternative DT visualizations used in Machine Learning. One of the major advantages of the PC-DT and SPC-DT methods is the ability to show the actual cases. It allows tracing of the misclassified cases to improve the accuracy of the DT model interactively.

SPC-DT also shows actual 2-D distributions in the ($X_i$,$X_j$) coordinates for adjacent nodes of the DT. For instance, when misclassified cases are close to the threshold of some nodes, then a user can interactively move the thresholds in these nodes in the SPC-DT or BC-DT visualization to change classification of those cases and see how it affects the classification of other training and validation cases.

In both SPC-DT and BC-DT methods, a user can interactively optimize the DT. While some other DT visualization such as [5] allow it by using 1-D distributions (Fig. 41), SPC-DT allows it in more detail in 2-D distributions. In addition, SPC-DT provides a more compact visualization of cases with a smaller number of tracing points. The BC-DT visualization has some similarities with visualization shown in Fig. 41 [5] where the coordinates are straight lines but does not trace actual cases from the root to the terminal nodes. In BC-DT, we use edges of the DT as bended coordinate lines. This makes the BC-DT more compact. It does not need drawing edges separately outside of the DT.

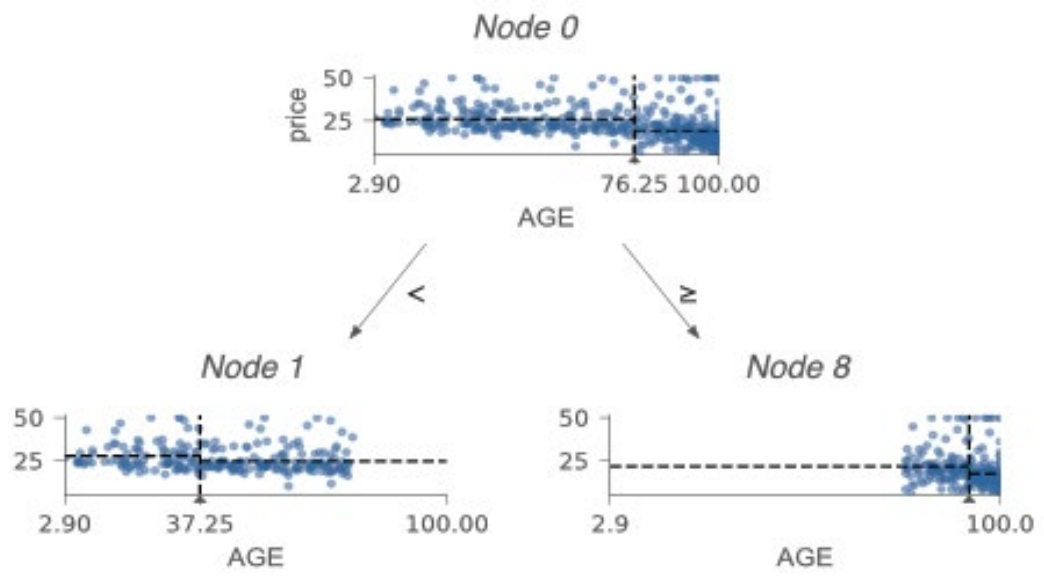

Fig. 41. Interactive splitting for DT nodes [5].

Commonly methods proposed to visualize decision trees visualize an individual “best” decision tree. This tree is selected automatically by DT algorithms among the trees with user selected desired characteristics: (1) accuracy, (2) minimum number of samples in the leaf, (3) height of the tree, and (4) the features to be included to the decision tree.

Many decision trees can have similar performance, but automatic methods **hide competitive alternatives** from the user. A "best" DT selected by some DT algorithm may not be the best for the user. A Timbertrek visualization system [30] allows a user to **select a DT interactively** from a set of almost-equally-optimal DT models known as the **Rashomon set** [23] motivated by the Rashomon effect from [24]. This is an important conceptual advantage of this system. It allows a user to see and explore much more DTs with the same characteristics (1)-(4) listed above. The **deficiency** of this system is that the user will be able to see only DTs selected by properties (1)-(4). Other properties can be also important for a user and can supersede some of these properties in their importance.

For example, a user can accept DTs with a lower number of cases in a few leaves, than the threshold that was set up as the minimum number of samples for all leaves, but the system will not show these DTs to the user. The same is applicable to the height of the tree, a user can allow a few longer branches than the height threshold. Next, a user can prefer a DT with a lower total accuracy, but higher accuracy for some class, e.g., for the malignant class to minimize the number of misclassified malignant cases as benign. These are just a few examples, but there are many such scenarios where a user may want some agency in the selection of the decision tree. Often a user can come up to such decisions in the process of the detailed analysis of the decision tree without explicit **requirements** formalized in advance, like (1)-(4). This is a **motivation** to the method that we are proposing to give a user a wider capability to analyze and build a desired DT.

The operations that can be conducted in BC-DT and/or SPC-DT system include the following interactive operations: (a1) adjusting thresholds in each node, (a2) joint adjusting and optimizing a pair of attributes (nodes of the DT), (a3) expanding DT, (a4) shrinking DT, (a5) blocking classifying some areas of the feature space, (a6) observing/tracing individual cases in the DT, (a7) simplifying visual form of the DT, (a8) visualizing a DT in two times more compact form than in other systems like [22, 23] by pairing nodes. Thus, our approach allows a domain expert to adjust DT to meet expert' needs. As we pointed out above these DTs can differ from selectable in [22].

Below we discuss the ways to combine advantages of SPC-DT and Timbertrek. A user can build and store several DTs in SPC-DT and then explore them. A user also can have access to a large set of DTs from Timbertrek, select, explore and modify them using SPC-DT tools. It can be done by (a1)-(a9) listed above. For instance, it can be done by creating the areas in SPC-DT, where the DT refuses to classify cases or allows to expand the DT. SPC-DT allows to create a knowledge base of these DTs with their characteristics doing search of most appropriate DTs like in [22] but with a wider set of search criteria. A user can work with BC-DT and SPC-DT systems on the regular basis, say, after the first session the user creates ten DTs of interest, during the next session the user can modify or combine DTs.

In the combined functionalities of SPC-DT and Timbertrek a user can select: (1) a set of features to include to the decision tree, (2) branches of the DT for exploration, (3) adjust thresholds for features at the tree's nodes, (4) explore how they fit user's needs, (5) observe branches and explore specific layers of the set of DTs, (6) split layers and branches, and others.

Both SPC-DT and Timbertrek empower domain experts and data scientists to easily explore many well-performing decision trees so they can find and collect those trees that best reflect their knowledge and values. Both visually summarize DTs based on the decision paths, enabling users to seamlessly transition across different components. Repositionable DTs with details of a decision tree and multiple windows allow users to compare and modify DTs to explore, enabling users to quickly identify decision trees with desired properties within capabilities of each tool.

Shifted paired Cardinals are used not only for decision trees but independently of the decision trees to find interpretable rules [19]. In those tasks, it is a significant challenge for Shifted Paired Coordinates to select an efficient pairing of coordinates that will allow for the discovery of patterns of cases of different classes clearly. Exploration of all possible pairings is not feasible, so a practical approach to avoid combinatorial exploration is using a decision tree to define the pairs of shifted paired coordinates as this paper illustrates with the SPC-DT approach.

The pairs are constructed using adjacent coordinates in the decision tree. In each branch of the DT, the coordinate that is at the root is paired with the coordinate in the second node of the branch. Next, the 3rd and 4th nodes form the second pair of coordinates, and so on. Some coordinates can be used in the DT several times with different thresholds. Respectively, the DT-based pairing will use them several times. This means that the decision tree approach allows us to expand the number of pairing alternatives beyond using each coordinate only in one pair.

Thus, the approach proposed in this paper allows (1) efficient visualization of the decision tree, (2) its improvement using this visualization, (3) communicating the DT to end-users, and (4) finding an efficient pairing of coordinates in SPC, which makes detecting patterns in the data much clearer. This opens an opportunity to use the SPC-DT approach beyond DTs.

### 5.3. Generalization of BC-DT and SPC-DT

Fig. 42 shows sharp and smooth versions of the BC-DT for their comparison. The selection of a particular form of BC-DT will depends on the user preferences. This can be a visually pleasing visualization, space used or others.

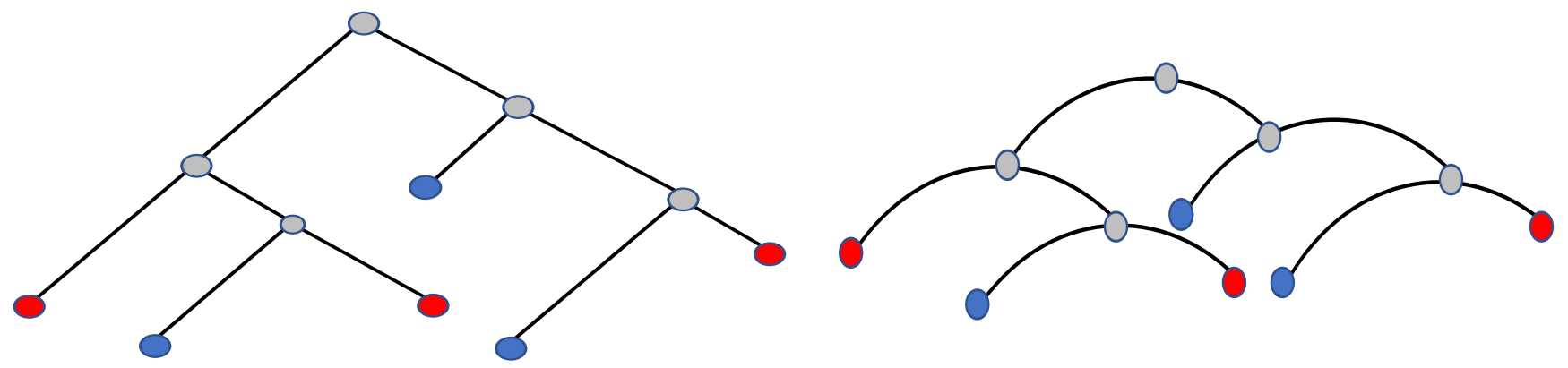

(a) Sharp bended DT.  (b) Smooth bended DT.

Fig. 42. Sharp and smooth bended DTs.

Fig. 43 shows the generalization of SPC-DT with bended smooth attributes. This approach links SPC-DT with smooth BC-DT shown above. It can be beneficial because some polylines can be simpler for analysis in curvilinear coordinates than in the straight coordinates.

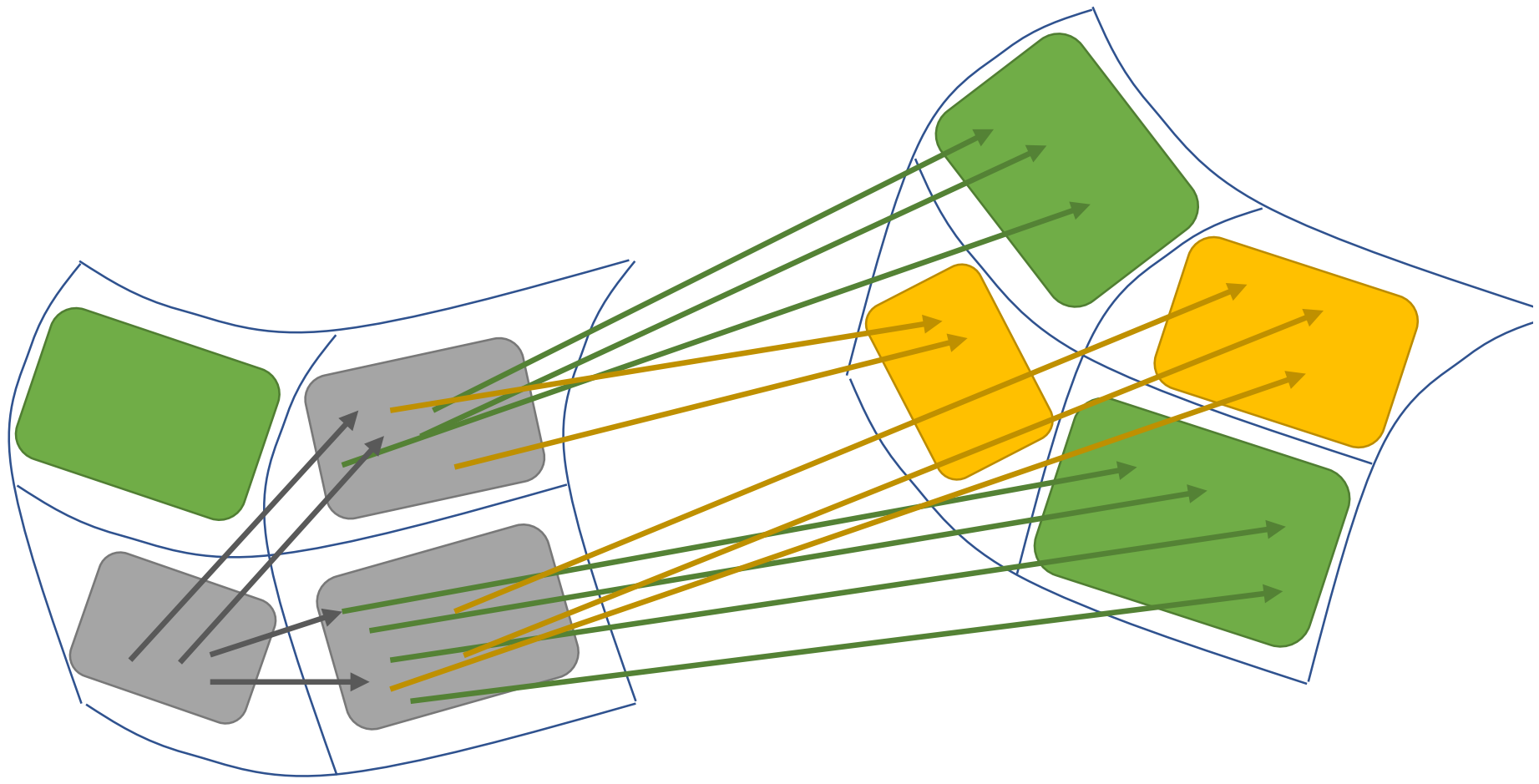

Fig. 43. SPC-DT with bended smooth attributes.

Fig. 44 illustrates it, with a curve in the traditional straight orthogonal coordinates on the left. In contrast, on the right, this curve became a straight horizontal preattentive line, because the X coordinates became a curve.

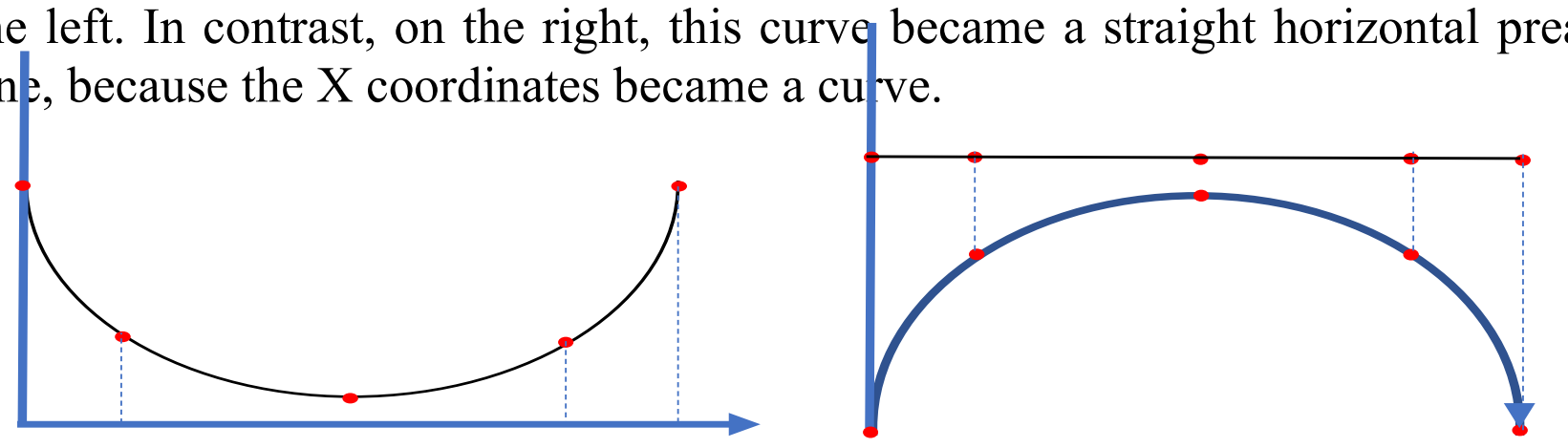

(a) Curve in Cartesian coordinates  (b) Curve transformed to a straight line in coordinates with one curvilinear coordinate.

Fig. 44. Transformation of a half circle to a straight line by transforming coordinates.

Similarly to presented above generalizations we can visualize DTs in other general line coordinates. It is already demonstrated in Section 2.1 with DT in parallel coordinates. With this approach methods to visulize DTs for other GLCs can be developed, including inline coordinates and n-gon coordiantes defined in [3].

### 5.4. Dealing with larger decsion trees

Dealing with larger decision trees requires additional features like zooming and panning and enhancement of the proposed methods. The current implementation allows to visualize DTs of the size like shown in the Fig. 45 for the Kaggle body performance dataset [30]. The problem is not only the size of the tree itself but the abilities to place the parts of the tree non-overlapping. This is a common issue for both SPC-DT and BC-DT. The current solution implemented for SPC-DT and BC-DT  is interactive. A user can drag parts of the decision tree to make the visualization more efficient. A future work can be to automate this process.

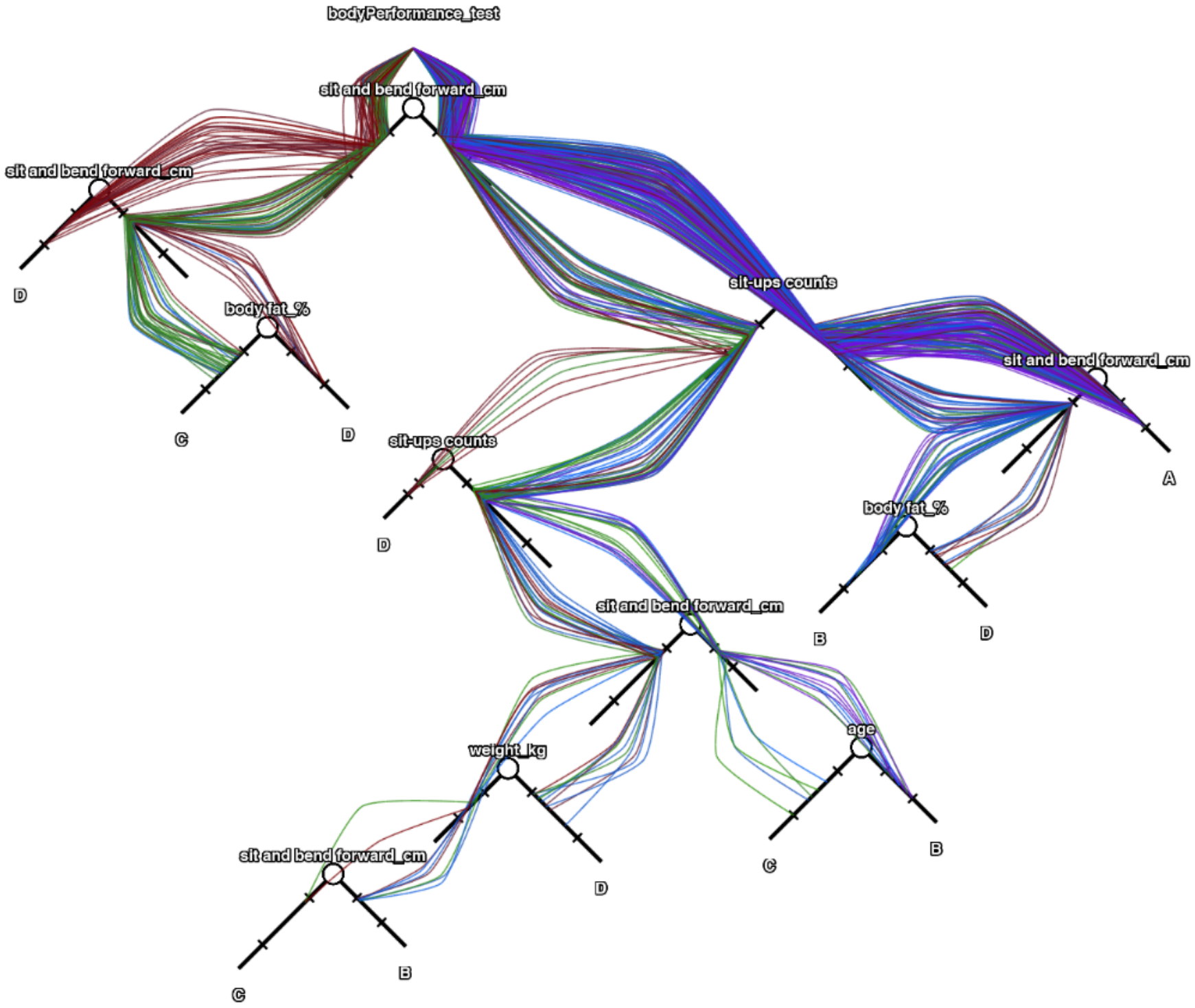

Fig. 45. Decsion treee for body performance data from [30].

## 6. Conclusions

This paper proposed two new DT visualization methods for machine learning. The features and advantages of these methods include showing relations between multiple pairs of attributes, in each plot, individual cases, closeness of the cases to the split thresholds, in the DT nodes, and the density of the cases, in the parts of the n-D space.

These features together help in getting confidence in the DT model interpretability and accuracy for the future predictions, avoiding overgeneralization and overfittings. The future work will be directed to be able to visualize the larger decision trees with more opportunities for user interaction. The developed experimental software is available at GitHub [29].

## References

[1] Dua, D. and Graff, C. UCI Machine Learning Repository, http://archive.ics.uci.edu/ml. Irvine, CA: University of California, School of Information and Computer Science, 2019.

[2] Elzen, van den, S. J., & Wijk, van, J. J. BaobabView: Interactive const-ruction and analysis of decision trees., VAST 2011, 151-160, IEEE, 2011,

[3] Kovalerchuk, B., Visual Knowledge Discovery and Machine Learning, Springer Nature, 2018.

[4] Ming Y, Qu H, Bertini E. RuleMatrix: visualizing and understanding classifiers with rules. IEEE transactions on visualization and computer graphics. 2018 Aug 20;25(1):342-52.

[5] Parr T., Grover P., How to visualize decision trees, https://explained.ai/decision-tree-viz/, 2019

[6] R2D3, A visual introduction to machine learning, http://www.r2d3.us/visual-intro-to-machine-learning-part-1/, 2019

[7] SAS: Working with Decision Trees, http://support.sas.com/documentation/cdl/en/vaug/68027/HTML/default/viewer.htm#n0q3i0zwng79kin1kb1zvpo9k312.htm, 2016.

[8] Schulz, H. Treevis.net: A tree visualization reference. IEEE Computer Graphics and Applications, 31(6):11–15, Nov 2011.

[9] Kovalerchuk B, Grishin V. Adjustable general line coordinates for visual knowledge discovery in n-D data. Information Visualization. 2019 18(1):3-2

[10] Kovalerchuk B, Grishin V. Reversible data visualization to support machine learning. In: International Conference on Human Interface and the Management of Information 2018, 45-59. Springer

[11] Scheibel W, Limberger D, Döllner J. Survey of treemap layout algorithms. In: Proceedings of the 13th International Symposium on Visual Information Communication and Interaction 2020, 1-9.

[12] Elkan C. The foundations of cost-sensitive learning. In: International joint conference on artificial intelligence 2001, Vol. 17, No. 1, 973-978.

[13] Rokach L, Maimon O. Classification trees. In: Data mining and knowledge discovery handbook 2009, 149-174. Springer, Boston, MA.

[14] Estivill-Castro V, Gilmore E, Hexel R. Human-in-the-loop construction of decision tree classifiers with parallel coordinates. In2020 IEEE International Conference on Systems, Man, and Cybernetics (SMC) 2020 Oct 11 (pp. 3852-3859). IEEE.

[15] Teoh ST, Ma KL. PaintingClass: interactive construction, visualization and exploration of decision trees. In: Proceedings of the ninth ACM SIGKDD international conference on Knowledge discovery and data mining 2003 Aug 24 (pp. 667-672).

[16] Kandogan. E., Visualizing Multi-Dimensional Clusters, Trends, and Outliers using Star Coordinates. Proc. ACM SIGKDD '01, pp. 107-116, 2001.

[17] A. Inselberg. Parallel Coordinates. Springer, 2009.
[18] Kovalerchuk B., Gharawi A., Decreasing Occlusion and Increasing Explanation in Interactive Visual Knowledge Discovery, In: S. Yamamoto and H. Mori (Eds.) Human Interface and the Management of Information. Interaction, Visualization, and Analytics, LNCS 10904, pp. 505–526, 2018, Springer, https://doi.org/10.1007/978-3-319-92043-6_42
[19] Wagle SN, Kovalerchuk B. Self-service Data Classification Using Interactive Visualization and Interpretable Machine Learning. In: Integrating Artificial Intelligence and Visualization for Visual Knowledge Discovery 2022 (pp. 101-139). Springer, Cham.
[20] McDonald R, Kovalerchuk B. Non-linear Visual Knowledge Discovery with Elliptic Paired Coordinates. In: Integrating Artificial Intelligence and Visualization for Visual Knowledge Discovery 2022 (pp. 141-172). Springer, Cham.
[21] Kovalerchuk, B., Phan J., Full interpretable machine learning in 2D with inline coordinates. 25th International Conference Information Visualisation IV-2021, Australia, Jul 5-9, 2021, Vol. 1, pp. 189-196, IEEE, DOI 10.1109/IV53921.2021.00038, arXiv:2106.07568.
[22] Wang ZJ, Zhong C, Xin R, Takagi T, Chen Z, Chau DH, Rudin C, Seltzer M. TimberTrek: Exploring and Curating Sparse Decision Trees with Interactive Visualization. In: 2022 IEEE Visualization and Visual Analytics (VIS) 2022 Oct 16 (pp. 60-64). IEEE.
[23] Xin R, Zhong C, Chen Z, Takagi T, Seltzer M, Rudin C. Exploring the whole rashomon set of sparse decision trees. arXiv preprint arXiv:2209.08040. 2022 Sep 16.
[24] Breiman L. Statistical modeling: The two cultures (with comments and a rejoinder by the author). Statistical Science, 16(3):199–231, 2001.
[25] Rudin C, Chen C, Chen Z, Huang H, Semenova L, Zhong C. Interpretable machine learning: Fundamental principles and 10 grand challenges. Statistics Surveys. 2022 Jan;16:1-85.
[26] Worland A., Wagle S., Kovalerchuk B., Visualization of Decision Trees based on General Line Coordinates to Support Explainable Models, in: 26th International Conference Information Visualisation, 2022, pp. 351–358, IEEE, arXiv:2205.04035.
[27] Rakotomalala R., Tanagra Software, http://eric.univ-lyon2.fr/~ricco/tanagra/en/tanagra.html
[28] Kovalerchuk B, Hayes D. Discovering Interpretable Machine Learning Models in Parallel Coordinates. In: 2021 25th International Conference Information Visualisation (IV) 2021 Jul 5 (pp. 181-188). IEEE, arXiv:2106.07474.
[29] GitHub: https://github.com/CWU-VKD-LAB., SPC-DT, Bended Attributes.
[30] Kaggle dataset: https://www.kaggle.com/datasets/kukuroo3/body-performance-data?resource=download
[31] Kovalerchuk B, Andonie R, Datia N, Nazemi K, Banissi E. Visual Knowledge Discovery with Artificial Intelligence: Challenges and Future Directions. In: Integrating Artificial Intelligence and Visualization for Visual Knowledge Discovery, 2022. pp. 1-27. Springer.